\documentclass[10pt,twocolumn,letterpaper]{article}

\usepackage[pagenumbers]{cvpr} % To force page numbers, e.g. for an arXiv version

\usepackage{amsmath}
\usepackage{wrapfig}
\usepackage{pifont}
\usepackage{makecell} 
\usepackage{diagbox}

\definecolor{cvprblue}{rgb}{0.21,0.49,0.74}
\usepackage[pagebackref,breaklinks,colorlinks,allcolors=cvprblue]{hyperref}

\def\paperID{6547} % *** Enter the Paper ID here
\def\confName{CVPR}
\def\confYear{2025}

\title{Seeing A 3D World in A Grain of Sand}

\author{Yufan Zhang$^{1}$ \quad Yu Ji$^{2}$ \quad Yu Guo$^{1}$ \quad Jinwei Ye$^{1}$\\
$^{1}$George Mason University \quad $^{2}$LightThought LLC\\
\small {\url{https://miniature-3dgs.github.io/}}
}

\begin{document}
\maketitle
\begin{abstract}

We present a snapshot imaging technique for recovering 3D surrounding views of miniature scenes. Due to their intricacy, miniature scenes with objects sized in millimeters are difficult to reconstruct, yet miniatures are common in life and their 3D digitalization is desirable. We design a catadioptric imaging system with a single camera and eight pairs of planar mirrors for snapshot 3D reconstruction from a dollhouse perspective. We place paired mirrors on nested pyramid surfaces for capturing surrounding multi-view images in a single shot. Our mirror design is customizable based on the size of the scene for optimized view coverage. We use the 3D Gaussian Splatting (3DGS) representation for scene reconstruction and novel view synthesis. We overcome the challenge posed by our sparse view input by integrating visual hull-derived depth constraint. Our method demonstrates state-of-the-art performance on a variety of synthetic and real miniature scenes.

\end{abstract}     
\section{Introduction}
\label{sec:intro}

\begin{verse}
\textit{To see a World in a Grain of Sand \\
And a Heaven in a Wild Flower, \\
Hold Infinity in the palm of your hand \\
And Eternity in an hour.\\
\hfill - William Blake}
\end{verse}

\begin{figure}[t]
    \centering
    \includegraphics[width=1\linewidth]{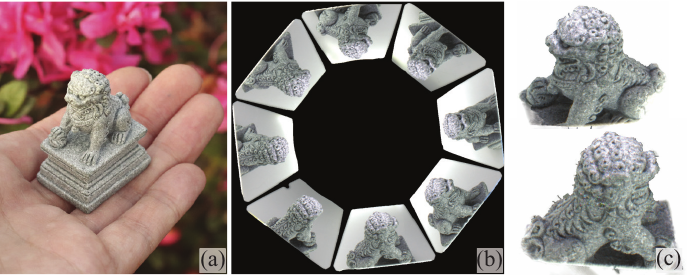}
    \caption{(a) A miniature scene with tiny object; (b) Example image captured by our catadioptric imaging system; (c) Novel view synthesis results.}
    \label{fig:teaser}
\end{figure}

Most existing works on 3D reconstruction or novel view synthesis focus on large-scale or life-size scenes. The tiny world of miniatures (\eg, objects sized in centimeters or even millimeters) is somewhat neglected. 3D reconstruction of miniature scenes is challenging due to limitations on lenses, image resolution and reconstruction accuracy. To take images of miniatures, one would need a macro lens to magnify the tiny objects, such that they could cover substantial amount of pixels in an image. Due to high magnification ratio, macro images usually have shallow depth of field, making it hard to capture all-in-focus image. In addition, many miniatures have little textures (as limited by their size), which poses challenge to photogrammetry-based 3D reconstruction. Yet miniatures scenes are common in life (see examples in Fig.~\ref{fig:teaser}): from toys and decorations in household to artisanal crafts and antiques in design studios and museums, high-quality 3D reconstruction not only opens up new ways for viewing and interacting with those miniatures, and also benefits their preservation~\cite{miniature} .

In this work, we present a snapshot solution for reconstructing miniature scenes and synthesizing their images in full surround 360$^\circ$ views. We design a novel planar catadioptric lens to enable zoom-in all-focus imaging and time-synchronized full surround multi-view acquisition. Specifically, we use the reflection between paired mirrors to fold the light path in order to zoom-in onto small object in short distance. We arrange eight pairs of mirrors on the sides of two nested octagonal pyramids to capture surrounding views of the scene in a single shot. We analyze the ray geometry of our lens and derive closed-form formulas for optimizing the mirror configuration based on the size of the scene, such that the multi-view coverage of the scene is optimal. We map our multi-view sub-images to virtual cameras and pre-calibrate the virtual camera parameters. The calibration only needs to run once when the mirror configuration is set. Since our calibrated camera parameters are highly accurate, they greatly benefit scene reconstruction. In contrast, self-calibration methods (\eg, structure from motion~\cite{schoenberger2016sfm}) heavily rely on scene features and are not reliable for miniature scenes with little textures.

We use the 3D Gaussian Splatting (3DGS)~\cite{kerbl3Dgaussians} for scene representation, and synthesize full surround novel views using our multi-view images and pre-calibrated camera parameters. Since our input views are sparse (\eg, 8 views on a 360$^\circ$ circle), we use additional depth constraints derived from visual hull to improve the 3DGS reconstruction. Specifically, we first calculate visual hull using the object's silhouette masks extracted from multi-view images. We then project the visual hull to depth map under each viewpoint. Since visual hull is a convex volume that fully enclose the object, we propose a weighted depth loss that penalizes more on depth values greater than the visual hull depth (which indicates the point is outside of the visual hull). Our depth loss is especially effective for surfaces with little textures, and it can be used for generic 3DGS when visual hulls are available.

We validate our method on both synthetic and real data. For synthetic experiments, we render images by emulating our imaging setup and quantitatively evaluate the novel view synthesis results. For real experiments, we custom-build lens prototypes using 3D-printed housing and acrylic mirrors, and capture images of a variety of miniature scenes. We compare our method with recent sparse-view 3DGS approaches, and demonstrate better rendering quality. Our main contributions are summarized as follows:
\begin{itemize}
\item We design a planar catadioptric imaging system for recovering 3D surrounding views of miniature scenes in a single shot.
\item We analyze the ray geometry of our catadioptric lens and derive closed-form formulas for calculating optimized lens parameters given scene information. 
\item We propose a novel weighted depth loss based on visual hull to improve 3DGS with sparse view input.
\item We build a prototype for the proposed imaging system and validate our approach on real miniature scenes. 
\end{itemize} 
\section{Related Work}
\label{sec:related}
Here we briefly review prior works that are most relevant to our imaging system and reconstruction algorithm. \\

\noindent\textbf{Mirror-based imaging system.}
Mirrors, either planar or curved, are widely used for building imaging systems with extended field-of-view or snapshot multi-view capacities. Imaging systems that involve both mirrors and refractive lenses are called catadioptric systems. They have been extensively studied and used for stereo~\cite{Wu2010DesignOS,Nene1998StereoWM,svoboda2002epipolar}, panoramic~\cite{chahl1997reflective,swaminathan2003framework}, surround-view~\cite{ahn2021kaleidoscopic,Lanman2007} and light field~\cite{Fuchs2013DesignAF, Ihrke2008FastIL, mukaigawa2011hemispherical} imaging. Reshetouski and Ihrke ~\cite{reshetouski2013mirrors} provide a comprehensive survey on the design and applications of various mirror-based imaging systems. Gluckman and Nayar~\cite{Gluckman00CVPR,Gluckman2001} show all possible configurations for catadioptric stereo and derive the reflection transformation for mirror-based image formation. Notably, kaleidoscopic imaging systems use the inter-reflection of mirrors to generate variations in viewpoints and illumination for 3D shape~\cite{wechsler2022kaleidomicroscope,reshetouski2011three} and reflectance~\cite{Ihrke2012,Han2003Measuring} reconstruction. Various techniques are explored for 3D reconstruction, including space carving~\cite{reshetouski2011three}, multi-view stereo~\cite{mas2019kaleidoscope}, structured light~\cite{ahn2021kaleidoscopic} and neural surface representation~\cite{Ahn2023CVPR}. Our work uses eight pairs of mirrors for full-surround imaging and the 3DGS framework for scene reconstruction. In contrast to kaleidoscopes, we avoid inter-reflection in our mirror system. Thus, our multi-view images are much easier to separate and analyze, although our angular resolution is sacrificed.\\ 

\noindent\textbf{3DGS with sparse view input.}
3D Gaussian Splatting (3DGS)~\cite{kerbl3Dgaussians} is an efficient radiance field representation that allows for high-quality real-time rendering of novel views, with relatively short training time. But the original 3DGS requires dense view supervision to achieve high quality rendering. With few reference views, the algorithm tends to overfit on inputs, resulting artifacts in unseen views. The problem gets even worse for surround-view rendering. Many recent techniques are proposed to allow robust 3DGS-based novel view synthesis when few views are available. A major trend is to use depth maps to provide additional supervision. FSGS~\cite{zhu2023FSGS} and SparseGS~\cite{xiong2023sparsegs} use monocular depth predicted by pre-trained model~\cite{Ranftl2022}. GS2Mesh~\cite{wolf2024gs2mesh} and InstantSplat~\cite{fan2024instantsplat} adopt depth learned from multi-view stereo. ReconFusion~\cite{wu2024reconfusion} and GaussianObject~\cite{yang2024gaussianobject} use diffusion model to synthesize addition views in order to provide dense viewpoint and depth supervision. RaDe-GS~\cite{zhang2024rade} improves the depth rasterization in 3DGS. In this work, we propose a new weighted depth loss that leverages depth map generated from visual hull to improve the quality of novel view synthesis with sparse surround-view input.

\section{Imaging System}
\label{sec:system}
In this section, we introduce our imaging system for snapshot full-surround miniature scene reconstruction. We first describe the design of our planar catadioptric lens, and then analyze the ray geometry of the lens and derive optimized lens parameters given scene information.

\subsection{Catadioptric Lens Design}
\begin{figure}[t]
    \centering
    \includegraphics[width=0.9\linewidth]{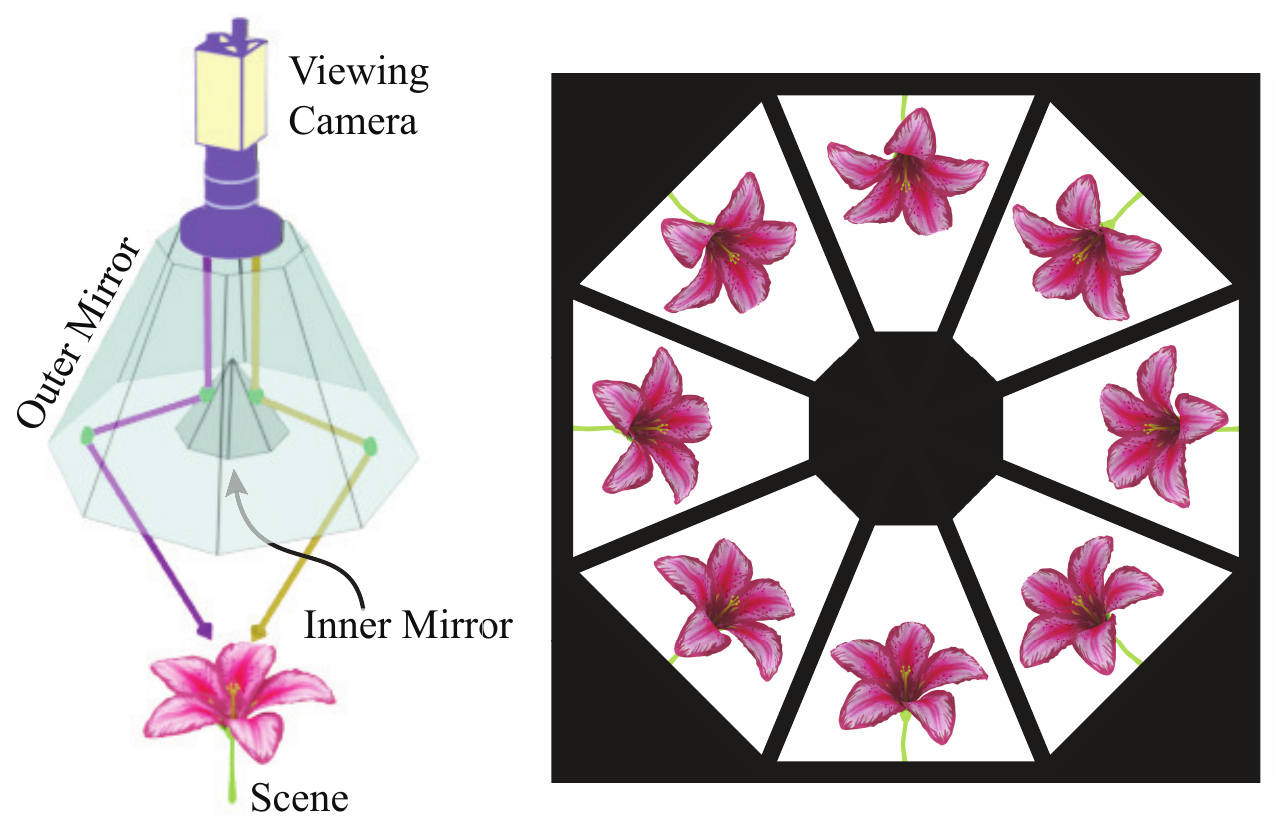}
    \caption{Left: A schematic illustration of our imaging system; Right: Sample image captured by the system.}
    \vspace{-8pt}
    \label{fig:lens}
\end{figure}

Our imaging system consists of a camera and a planar catadioptric lens that allows for synchronized full-surround multi-view image acquisition. The conceptual design is illustrated in Fig.~\ref{fig:lens}. Our catadioptric lens is made with eight pairs of flat mirrors that are circularly arranged on the sides of two nested octagonal pyramids. In each pair, the mirrors are facing each other with different tilting angles and shifted locations. Through two times of reflections, the mirror pair guides light from a scene that is underneath to the viewing camera on top. In this way, the light paths are folded such that we are able to image small objects in short distance, as if they are imaged by a zoom lens. Similar idea on light path folding is used in Folded Optics (or Origami Optics)~\cite{Tremblay07,Tremblay2009UltrathinFI} for compact wide-angle lens design.  

Then we arrange eight such mirror pairs along a circle, on surfaces of two nested octagonal pyramids, in order to capture full-surround multi-view images. A sample image captured by our imaging system is shown in Fig.~\ref{fig:lens}. Each sub-image is formed by a pair of mirrors. With eight mirror pairs, we obtain eight sub-images that surround the scene in 360$^\circ$ and are time-synchronized by optics. Note that we purposely arrange the mirrors such that light only bounces once on a piece of mirror. Therefore, there is no inter-reflection in our sub-images. Similar design with mirror pairs is studied for stereo image acquisition~\cite{Kim2006}. But two mirror pairs on opposite sides have non-overlapping views and is not practical for stereo vision, whereas our circular arrangement guarantees overlapping in neighboring views and provides full-surround coverage of the scene.\\

\begin{figure}[t]
    \centering
    \includegraphics[width=1\linewidth]{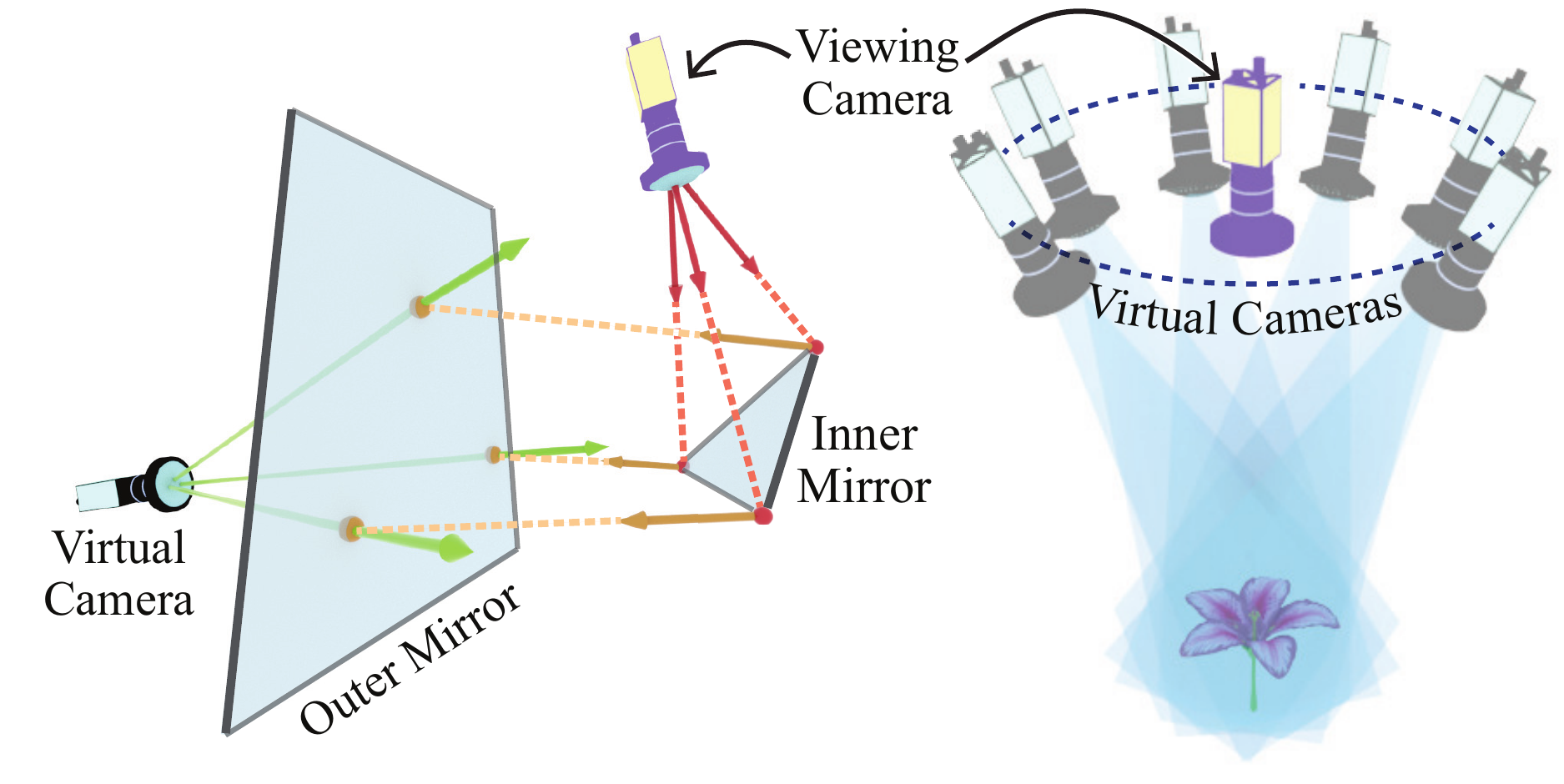}
    \caption{We map mirror reflections to virtual camera views by unfolding the light paths.}
    \vspace{-8pt}
    \label{fig:virtual}
\end{figure}

\noindent\textbf{Mapping to virtual cameras.} By unfolding the reflection light paths, we are able to map our sub-images to virtual camera views (as if there are no mirrors). This unfolding process and mapped virtual cameras are illustrated in Fig.~\ref{fig:virtual}. The virtual cameras are evenly distributed on a circle, looking inward to the scene in the center. All cameras share the same intrinsic parameters, since mirror reflection linearly maps an image. The camera poses are determined by the mirror parameters (\eg, tilting angles and positions). By varying the mirror configuration, we are able to adjust the viewing directions and baseline of surrounding views, so as to optimize their coverage of the scene. In Sec.~\ref{sec:georay}, we analyze the multiview coverage in relationship to the mirror configuration in details. \\

\noindent\textbf{Comparison with kaleidoscopic imaging.} Kaleidoscopic imaging systems use the inter-reflection of mirror chambers for snapshot surrounding view acquisition \cite{Ihrke2012,ahn2021kaleidoscopic}. One key distinction between our design and kaleidoscopes is that we only allow one-bounce reflection on mirrors and so we do not have inter-reflection in images. With simpler ray geometry, our system is much easier to calibrate and we mitigate the challenging ``labeling" problem on determining the mirror sequence involved in inter-reflections. However, the angular resolution is traded off in our design: we have fewer viewing directions comparing to kaleidoscope since the inter-reflected views are absent. Nevertheless, we show in Sec.~\ref{sec:method} and through experiments that our sparse surrounding views are sufficient for 3DGS-based reconstruction and full surround novel view synthesis. 

\subsection{Ray Geometry Analysis}
\label{sec:georay}
Next we analyze the light transport inside of our lens and derive the relationship between our viewing volume and mirror parameters. We then use the formulas to optimize the mirror configuration based on the scene dimensions, such that the multi-view coverage of the scene is optimal.\\

\noindent\textbf{Effective viewing volume.} The effective viewing volume of our imaging system is the intersection of viewing frustums of all virtual cameras. Here we derive the relationship between the dimension of effective viewing volume and the mirror parameters. 

\begin{figure}[t]
    \centering
    \includegraphics[width=1\linewidth]{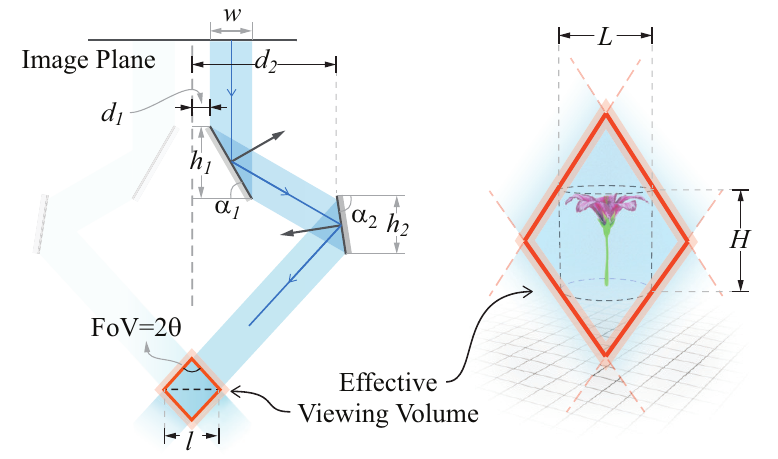}
    \caption{The reflection light path of our mirror pair and the optimal coverage of a scene by our effective viewing volume.}
    \label{fig:lightpath}
\end{figure}

Since our sub-images exhibit weak perspective with long effective focal length, we assume orthographic camera model for simplicity. The effective viewing volume is carved out by eight beams of parallel light as a polyhedron that can be approximated as a symmetric double pyramid. We characterize this volume by its side angle at apex ($\theta$) and the widest length at the base ($l$). WLOG, we illustrate the formation of effective viewing volume in a 2D cross-section (see Fig.~\ref{fig:lightpath}), since our mirror design is symmetric. We parameterize the mirrors in a pair using their tilting angles ($\alpha_{1,2}$), vertically projected heights ($h_{1,2}$), and distances  ($d_{1,2}$) between their upper edges and the central ray (\ie, vertical ray from the imager center). We denote the inner mirror as $M_1$ and the outer one as $M_2$. Subscript of the parameters indicate their correspondence to the mirrors. 

Since light path is reversible, we trace rays from the camera to the scene. Given a vertical ray from camera, it is first bounced on $M_1$ and then on $M_2$ before reaching the scene. The ray's incident angle to $M_1$ (\ie, the angle between the incident ray and the mirror's normal) is: $\omega_1 = \alpha_1$. It's incident angle to $M_2$ can then be calculated as: $\omega_2 = 2\alpha_1- \alpha_2$. Since rays reflected from $M_2$ intersect with the symmetric beam from the opposite side to form the effective viewing volume, the pyramid's side angle at base can be calculated as: $\beta = 90^\circ - \alpha_2 + \omega_2 = 90^\circ - 2(\alpha_2 - \alpha_1)$. The side angle at apex is thus: 
   \begin{equation}
   \label{eq:angle}
   \theta = 90^\circ - \beta = 2(\alpha_2 - \alpha_1).
   \end{equation}

Let $\Delta\alpha = \alpha_2 - \alpha_1$. We define $2\theta = 4\Delta\alpha$ as the field-of-view (FoV) of our catadioptric lens. Detailed derivation of these angles can be found in the supplementary material. 

The width of parallel light beam that can enter the camera is determined by $M_1$'s size and angle: $w =h_1 / \tan\alpha_1$. Since $w$ remains constant after reflection, we can calculate the base length $l$ of the viewing volume as: 
   \begin{equation}
   \label{eq:length}
   l = \frac{w} {\cos\theta} = \frac{h_1}{\tan\alpha_1\cdot\cos2\Delta\alpha}.
   \end{equation}
\noindent\textbf{Discussions on mirror design.} From Eqs.~\ref{eq:angle} and~\ref{eq:length}, we can see that the dimension of effective viewing volume is determined by the angle and size of $M_1$ ($h_1$ and $\alpha_1$) and the angle difference between $M_1$ and $M_2$ ($\Delta\alpha$). The greater the $\Delta\alpha$, the larger the FoV. In order to form reflection images from underneath without inter-reflection, the mirror parameters need to satisfy the following three conditions: 
\vspace{4pt}
\begin{enumerate}[label=(\roman*)]
\item $45^\circ < \alpha_1 < \alpha_2 < 90^\circ$, such that light from scene underneath can be reflected to the imager on top;
\item $h_2 \geq \dfrac{\sin\alpha_2}{\tan\alpha_1 \cdot\cos(\alpha_2 - 2\alpha_1)} \cdot h_1,$
such that $M_2$ can cover the entire light beam reflected from $M_1$;
\item $d_2 \geq \dfrac{\tan \alpha_1 + \cot2\Delta\alpha}{\tan \alpha_1\cdot(\cot2\Delta\alpha - \cot 2\alpha_1)} \cdot{h_1} + d_1,$
such that no inter-reflection occurs between $M_1$ and $M_2$.
\end{enumerate}
\vspace{4pt}
Please see the supplementary material for derivation of these conditions. Note that although the size and location of $M_2$ ($h_2$ and $d_2$) would not affect the dimension of effective viewing volume dimension, changing these parameters would result in the viewing volume to be shifted vertically. It is preferable for $d_2$ to take smaller values, since the resulting viewing volume would be closer to the camera. \\

\noindent\textbf{Optimized mirror configuration.} 
We then show how to find optimal configuration for the mirror pair, given the size of the scene. Since mirror parameters determine the dimension of effective viewing volume. We first find a viewing volume that provides the optimal scene coverage, and then use Eqs.~\ref{eq:angle} and~\ref{eq:length}, along with the three conditions to calculate mirror parameters. We use the following two criteria to find the optimal viewing volume: 1) in order to have a complete reconstruction of the scene, the viewing volume should be large enough to fully enclose the scene; 2) larger FoV is preferred in order to have more coverage on the lateral sides. Since FoV is intersected by the viewing directions of virtual cameras, the larger the FoV, the more oblique the viewing angles are, and thus the more coverage on the side.  

Since the amount of light that can be received by the imager is bounded by its size, we first allow the width of the parallel beam $w$ to take its maximum value, which is half of the sensor size\footnote{We assume orthographic model. Under perspective model, this value also depends on the lens.}, and we denote it as $w_{\text{max}}$. Since $w = h_1 / \tan\alpha_1$, we can adjust the angle and length of $M_1$ to achieve $w_{\text{max}}$. Given $w_{\text{max}}$, the dimension of effective viewing volume is only related to $\Delta\alpha$. The greater the $\Delta\alpha$, the larger the FoV and base length $l$, but the smaller the vertical height $h = w_{\text{max}}/\sin2\Delta\alpha$. We then set out to find the viewing volume with the largest FoV that can fully enclose a scene with known size. We approximate a scene using its bounding box with size $W\times L \times H$ (where $H$ is the vertical height and we assume $L \geq W$). The FoV 
takes its largest value when the scene's bounding box is inscribed in the viewing volume (see Fig.~\ref{fig:lightpath}). This is because when FoV further increases beyond this value, the vertical height $h$ decreases, which would result in the scene being partially outside of the viewing volume. 

Next we derive the FoV value under this circumstance. With similar triangles, we have $(h - H)/h = L/l$. By substituting $h = w_{\text{max}}/\sin2\Delta\alpha$ and $l = w_{\text{max}}/\cos2\Delta\alpha$, we have $H\cdot\sin2\Delta\alpha + L\cdot\cos2\Delta\alpha =  w_{\text{max}}$. By applying the sine angle addition identity, we have:
\begin{equation}
\label{eq:opt_angle}
    \Delta\alpha = \frac{1}{2}(\arcsin(\frac{w_{\text{max}}}{\sqrt{L^2 + H^2}}) - \arctan(\frac{L}{H})).
\end{equation}
The largest FoV is $4\Delta\alpha$, which can be calculated given the scene size. Eq.~\ref{eq:opt_angle} indicates the angle difference between $M_1$ and $M_2$, which is the optimal mirror configuration to achieve the largest FoV.

\section{Scene Reconstruction}
\label{sec:method}
\begin{figure}[t]
    \centering
    \includegraphics[width=1\linewidth]{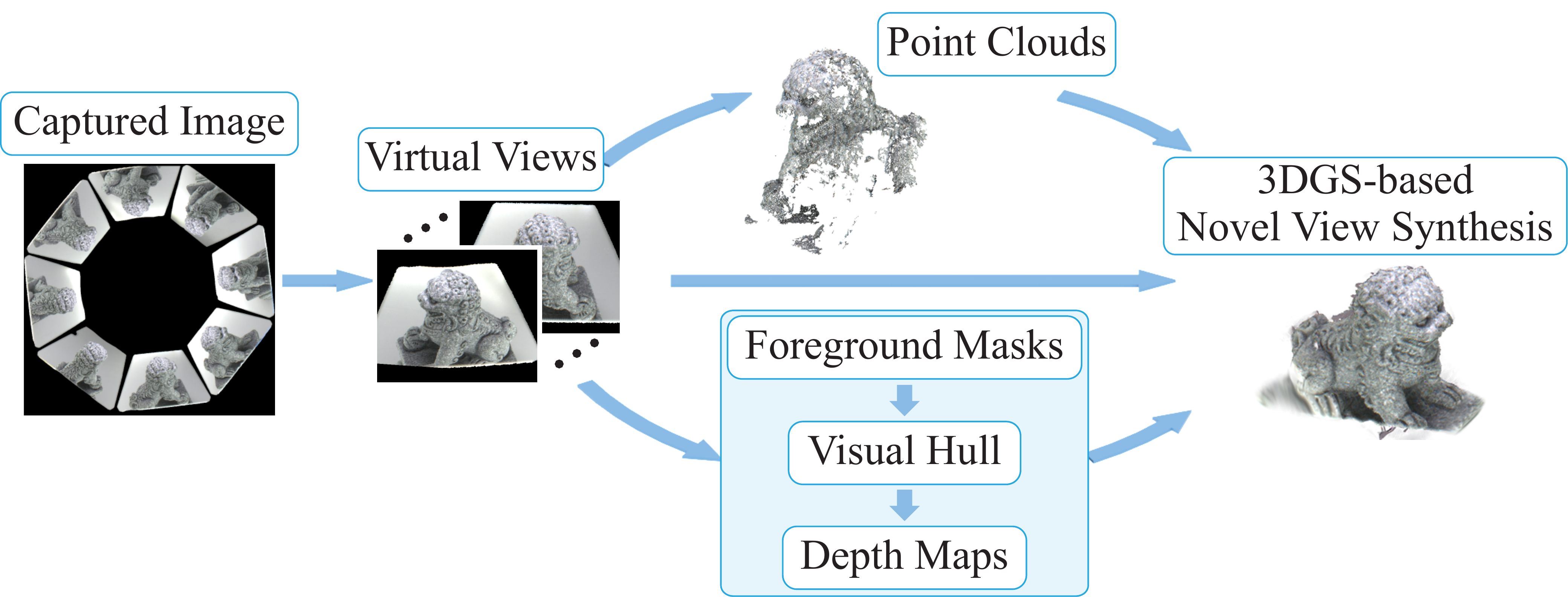}
    \caption{Our overall algorithmic pipeline.}
    \label{fig:pipeline}
\end{figure}

In this section, we show how to use our capture image for scene reconstruction and novel view synthesis. Specifically, we adopt the 3D Gaussian Splatting (3DGS) framework. We use depth supervision constrained by visual hull to improve the quality of full-surround novel view synthesis with sparse reference views. The overall algorithmic pipeline of our method is shown in Fig.~\ref{fig:pipeline}.%present our 3D Gaussian Splatting (3DGS) based method for scene reconstruction and novel view synthesis. 

\subsection{Pre-processing Steps}
Since our multi-view images are combined into a single image, we first segment our captured image into individual multi-view images. We obtain the segmentation mask by taking one image of a solid color background. By applying the mask, we obtain eight images as if captured by surrounding virtual cameras. Since the orientation of virtual cameras are flipped on opposite sides, we re-project the multi-view images onto new viewing planes, such that the transition of view poses is smooth on a circle. We also create foreground masks to segment out the object of interest. This can be done using off-the-shelf segmentation tools, such as Segment Anything~\cite{kirillov2023segany}. Please see examples of our input and processed images in the supplementary material. 

We also pre-calibrate our imaging system using a small checkerboard target to obtain the intrinsic and extrinsic parameters of virtual cameras. The extrinsic parameters are re-calculated after view re-projection. The calibration process only needs to run once when our catadioptric lens is fixed on the camera. Pre-calibrating these parameters greatly benefit miniature scene reconstruction, since these tiny scenes often lack textures, and this would result in self-calibration methods (\eg, COLMAP~\cite{schoenberger2016sfm}) to fail. With the camera parameters, we estimate an initial point cloud using structure from motion. We then use the eight multi-view images, along with the camera parameters and the initial point cloud as input for 3DGS-based scene reconstruction.

\subsection{3DGS Representation}
3DGS, first introduced by Kerbl \etal~\cite{kerbl3Dgaussians}, represents a scene using a set of 3D Gaussian elements. It is a more compact and efficient 3D representation, comparing to NeRF~\cite{mildenhall2020nerf} and classical triangle meshes. It has been demonstrated great success in many rendering applications~\cite{tang2023dreamgaussian,zhou2024drivinggaussian,kocabas2024hugs}.

A 3D Gaussian is defined by its center position \(\boldsymbol{\mu} \in \mathbb{R}^3\) and a covariance matrix \(\Sigma \in \mathbb{R}^{3 \times 3}\):
\begin{equation}
G(\boldsymbol{x}) = e^{-\frac{1}{2}(\boldsymbol{x} - \boldsymbol{\mu})^\top \Sigma^{-1} (\boldsymbol{x} - \boldsymbol{\mu})},
\end{equation}
where $\boldsymbol{x}$ is a point on the Gaussian, and $\Sigma$ can be further decomposed into rotation and scaling matrices ($R$ and $S$ respectively): $\Sigma = RS S^\top R^\top$. 

To render an image, the 3D Gaussians are projected by: $\Sigma' = JW\Sigma W^\top J^\top$, where $W$ is the viewing transformation matrix and $J$ is the Jacobian of the affine approximation of the projective transformation. A pixel's color $\mathbf{C}$ is calculated by blending the color of ordered Gaussians that overlap the pixel, similar to the NeRF-style rendering: 
\begin{equation}
\label{eq:GS_color}
\small
\mathbf{C} = \sum_{i=1}^{n} c_i \alpha_i T_i,
\end{equation}
where $c_i$ is the color of the Gaussian, \(\alpha_i\) is the opacity of the 2D projected Gaussian and $T_i = \prod_{j=1}^{i-1} (1 - \alpha_j)$ is the transmittance along the ray. 

\subsection{Optimization}
To optimize the Gaussian parameters, we compare the rendered output against our input reference views using the following loss function: 
\begin{equation}
\small
\mathcal{L} = \lambda_1 \mathcal{L}_1 + \lambda_2 \mathcal{L}_{\text{D-SSIM}} + \lambda_3 \mathcal{L}_{\text{depth}},
\end{equation}
where $\lambda_{1,2,3}$ are weighting factors for balancing the terms. We use $\lambda_1 = 0.8$, $\lambda_1 = 0.2$ and $\lambda_3 = 0.5$ in our experiments. \(\mathcal{L}_1\) and \(\mathcal{L}_{\text{D-SSIM}}\) are standard losses that evaluate the color similarity between the rendered image and the reference image. Since our input views are sparse with limited overlapping, using the color losses alone results in severe artifacts when synthesizing unseen views. We therefore add the third loss $\mathcal{L}_{\text{depth}}$ (which will be described in the following paragraph), a depth loss constrained by visual hull-generated depth maps, to improve the rendering quality of novel views.\\ 

\noindent\textbf{Visual hull-constrained depth loss.} By using the camera parameters and foreground object masks (which indicate their sillouettes), we can carve out a visual hull~\cite{laurentini1994visual} for objects in the scene. We then project the visual hull under each reference view and render depth map $\mathbf{D}_{\text{VH}}$ for depth regularization.

We use the rasterizer provided by \cite{hierarchicalgaussians24} to render depth map $\mathbf{D}_{\text{render}}$ from 3D Gaussians. Specifically, the depth is rendered by replacing the color $c_i$ in Eq.~\ref{eq:GS_color} with the depth $d_i$ of the Gaussian's center: $\mathbf{D}_{\text{render}} = \sum_{i=1}^{n} d_i \alpha_i T_i$. Since $\mathbf{D}_{\text{render}}$ and $\mathbf{D}_{\text{VH}}$ have consistent scale, we calculate the depth loss $\mathcal{L}_{\text{depth}}$ as a weighted L1 norm: 
\begin{equation}
\mathcal{L}_{\text{depth}}  = \frac{2}{1+e^{\Delta d_i}} |\mathbf{D}_{\text{render}} - \mathbf{D}_{\text{VH}}|,
\end{equation}
where $\Delta d_i = \mathbf{D}_{\text{render}} (\boldsymbol{p}_i) - \mathbf{D}_{\text{VH}}(\boldsymbol{p}_i)$, with $\boldsymbol{p}_i$ refers to a pixel in image.

\begin{wrapfigure}{l}{0.19\textwidth} 
    \centering
    \includegraphics[width=0.19\textwidth]{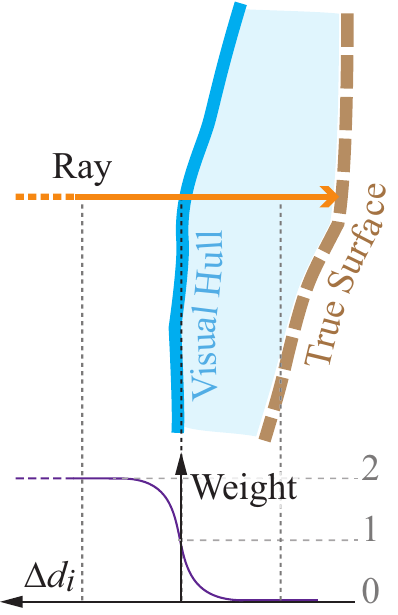}
    \caption{S-shape weight.}
    \label{fig:weight}
\end{wrapfigure}

\noindent $\mathcal{L}_{\text{depth}}$ is weighted by a s-shaped logistic function with values greater than 1 when $\Delta d_i > 0$ (see Fig.~\ref{fig:weight}). We introduce this weight based on the observation that visual hull is a convex enclosure of the actual geometry. Therefore, we use larger weight for points that are outside of the visual hull (\ie, $\Delta d_i > 0$). For points inside of the visual hull, we gradually decrease the weight to zero, since its depth may still be correct even though $\Delta d_i \neq 0$, if the surface is concave. This weight allows us to model the visual hull-constrained depth regularization more precisely.
\section{Experiments}
\label{sec:experiment}
We evaluate our method on both synthetic and real-captured data. Our codes and input data will be made available on our project website.

\subsection{Experimental Setup}
\noindent\textbf{Synthetic data generation.}
We simulate synthetic images as captured by our proposed catadioptric imaging system using Autodesk 3ds Max. We build 3D models of mirror pairs and render image using a viewing on top of the mirror pyramid. We render 7 scenes with various levels of complexity in terms of texture and geometry. We render ground truth masks for segmenting the multi-view images and foreground objects. In order to calibrate the virtual cameras, we render checkerboard images with different poses. We use ground truth intrinsic parameters and only calibration the extrinsic parameters. We also render ground truth novel view images for quantitative evaluation. Since our 8 virtual cameras are arranged on a circle (which are our input reference views), we render 24 views along the same circle and use them as ground truth for evaluating novel view synthesis results.\\ 

\begin{figure}[t]
    \centering
    \includegraphics[width=1\linewidth]{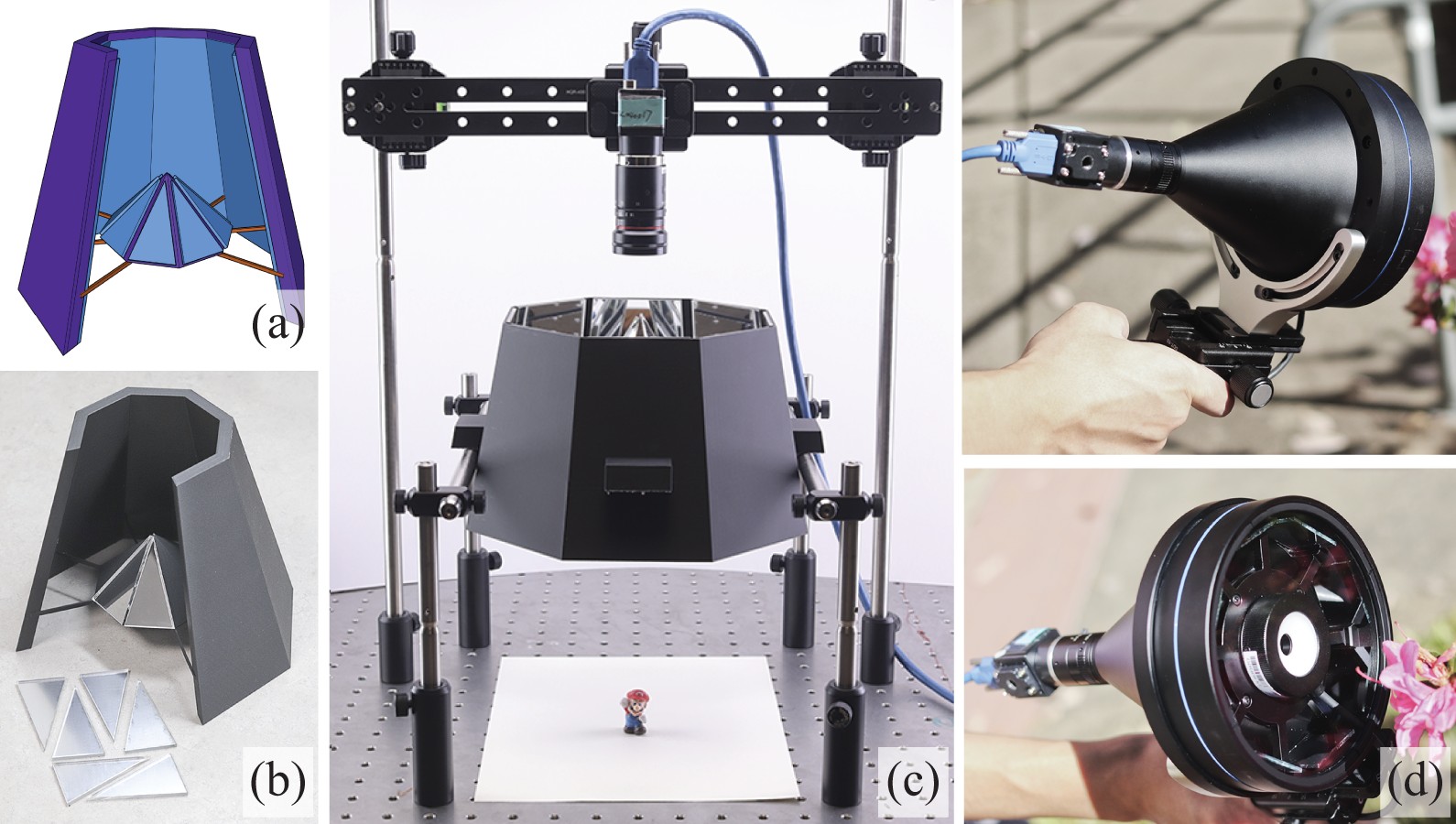}
    \caption{Prototypes of our mirror lens. (a) 3D model of our lens housing; (b) 3D printed housing with acrylic mirror pairs adhered to the inside; (c) Our imaging setup; (d) Our custom-made portable prototype.}
    \vspace{-12pt}
    \label{fig:system_hardware}
\end{figure}

\noindent\textbf{Real experiment setup.} We build prototypes for our proposed mirror lens in order to perform real experiments. In order to validate our derivation on mirror angle optimization, we custom-build lens with different mirror angles using 3D-printed housing and acrylic mirrors. The lens 3D model and prototype components are shown in Fig.~\ref{fig:system_hardware} (a) and (b). We mount a 5 megapixel FLIR camera with 16mm lens on top of the lens to capture images of a scene that is place below the mirror lens. Since our lens block light from above the scene, we mount LED light underneath the inner pyramid in order for the scene to be well lit. We try lenses with different mirror angles and spacing. Effects of these mirror configuration changes with respect to scene coverage are consistent with our derivations in Sec.~\ref{sec:georay}. Please see the supplementary material for example images captured with different mirror configurations.

We also order a customized mirror lens with sharping imaging quality from a lens maker (see Fig.~\ref{fig:system_hardware} (d)). We use this lens for capturing real scene data. We use the same 5 megapixel FLIR camera with 12mm lens for image acquisition. Since the lens is compact and portable, we capture miniature scenes from both indoor and outdoor. In total, we capture 10 indoor scenes and 2 outdoor scenes. The size of miniature objects in our scene is in the range of 1cm to 5cm. 

We pre-calibrate our imaging system to obtain intrinsic and extrinsic parameters for the virtual cameras. Since mirror reflection doesn't change intrinsic parameters, we first estimate the intrinsics with all available views and use the same set of parameters for all virtual cameras. We then calibrate the extrinsics after mounting the mirror lens. We also run bundle adjustment to further refine the camera poses. The final re-projection error of our calibration is 0.77.

\begin{figure*}[t]
    \centering
    \includegraphics[width=0.98\linewidth]{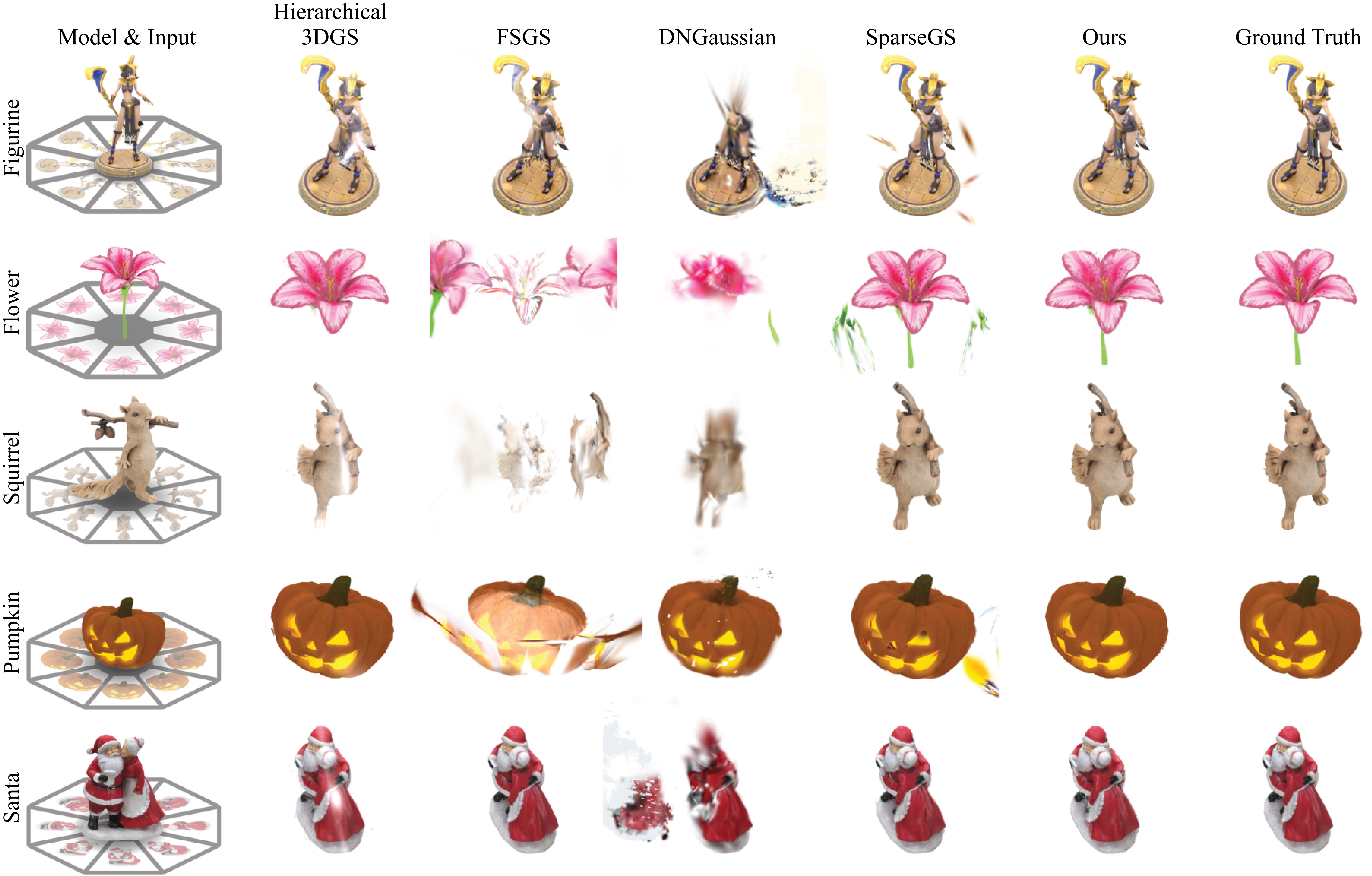}
    \caption{Visual comparison results on synthetic data. See supplementary material for more results and videos of full-surround rendering.}
    \vspace{-4pt}
    \label{fig:sync_res}
\end{figure*}

\noindent\textbf{Code implementation.} We implement the pre-processing steps, including image re-projection, visual hull generation and visual hull depth rendering \etc., using MATLAB. With our multi-view images and calibrated camera parameters, we use Agisoft Metashape\footnote{https://github.com/agisoft-llc} to generate an initial point cloud. We then convert the point cloud and camera parameters to COLMAP format, so they can be used by 3DGS. We use the latest Hierarchical 3DGS~\cite{hierarchicalgaussians24} as our backbone code. We integrate our visual hull-based depth regularization and optimize the rendering using foreground object mask, such that only the foreground is used for optimization. All our code are run on a computer with NVIDIA 4090 GPU. The average running for reconstructing a scene using 8 reference views, each with resolution $800 \times 800$, is around 2 minutes.

\begin{table}[t]
\caption{Quantitative comparison results.}
\centering
\begin{tabular}{c !{\vrule width 1pt} c c c}\Xhline{1pt} 
\textbf{Method} & \textbf{SSIM $\uparrow$} & \textbf{PSNR $\uparrow$} & \textbf{LPIPS $\downarrow$} \\
\Xhline{1pt}
Hierarchical 3DGS & 0.9750 & 26.8259 & 0.0298 \\
\hline
FSGS & 0.7844 & 18.9281 & 0.1100 \\
\hline
DNGaussian & 0.9128 & 21.3979 & 0.1296 \\
\hline
SparseGS & 0.9756 & 31.8415 & 0.0367\\
\hline
\textbf{Ours} & \textbf{0.9783} & \textbf{32.4792} & \textbf{0.0265} \\
\Xhline{1pt}
\end{tabular}
\vspace{-8pt}
\label{tab:comparison}
\end{table}

\subsection{Synthetic Results}
We validate our approach on our simulated data, and compare our results with recent state-of-the-art 3DGS algorithms: Hierarchical 3DGS~\cite{hierarchicalgaussians24}, FSGS~\cite{zhu2023FSGS}, DNGaussian~\cite{li2024dngaussian} and SparseGS~\cite{xiong2023sparsegs}. Most of these methods are optimized for taking sparse view input. It is worth noting that since our scenes are small and lack distinct features, COLMAP fails to run on all of our scenes. For fair comparison, we use our pre-calibrated camera parameters and initial point cloud as input for all compared algorithms. For each scene, we use 8 reference views as input and render 24 novel views. Visual comparison results on five scenes are shown in Fig.~\ref{fig:sync_res}. Quantitative comparison results using standard metrics (\eg, SSIM, PSNR and LPIPS) are shown in Table~\ref{tab:comparison}. The metric values are averaged over all scenes and all rendered novel views.

We can see that our method outperforms the state-of-the-arts in both qualitative and quantitative comparisons. The other sparse view methods do not perform very well. It is likely because that the monocular depth map they use is less accurate than our visual hull depth (see supplementary material for an ablation study on depth loss).

\begin{figure*}[t]
    \centering
    \includegraphics[width=0.98\linewidth]{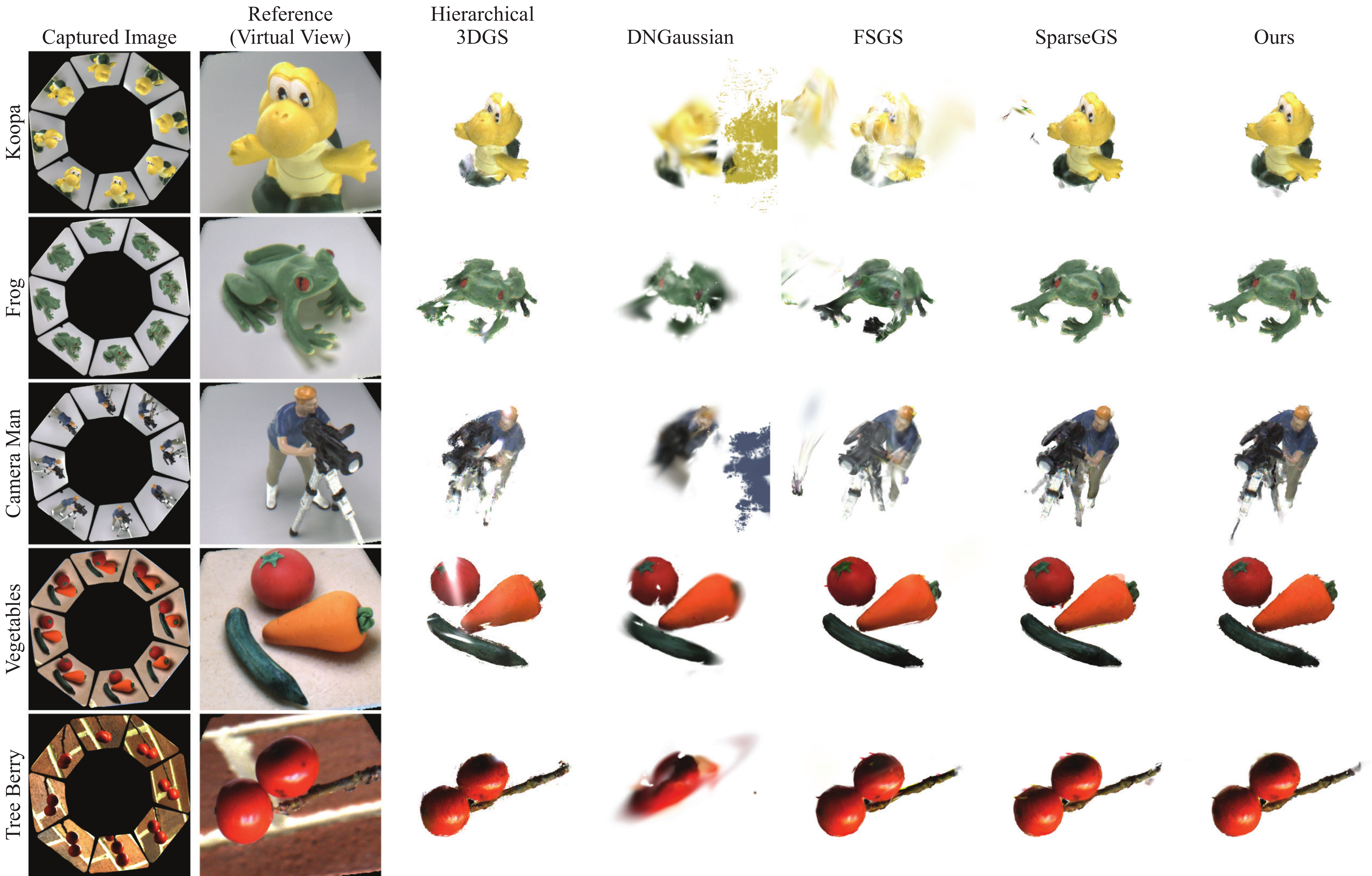}
    \caption{Visual comparison results on real data. See supplementary material for more results and videos of full-surround rendering.}
    \vspace{-4pt}
    \label{fig:real_res}
\end{figure*}

\subsection{Real Results}
We perform comparisons on real scene images taken with our mirror lens. Fig.~\ref{fig:real_res} show visual comparison results against the state-of-the-arts. We also show our captured image (processed with the multi-view mask) and one re-projected virtual view, which is used as input reference image for 3DGS (we use eight reference views in total). All the scenes are composed with miniature objects with size between 1cm and 4cm. The ``tree berry" scene is captured outdoor with our portable lens. Same as the synthetic experiments, all compared methods take our pre-calibrated camera parameters and initial point cloud, since COLMAP fails on our scenes. The ``camera man" scene is challenging with thin structures around 1mm (\eg, the tripod legs). Our method is able to recover the thin legs, whereas all other methods cannot. 

\begin{figure}[t]
    \centering
    \includegraphics[width=0.95\linewidth]{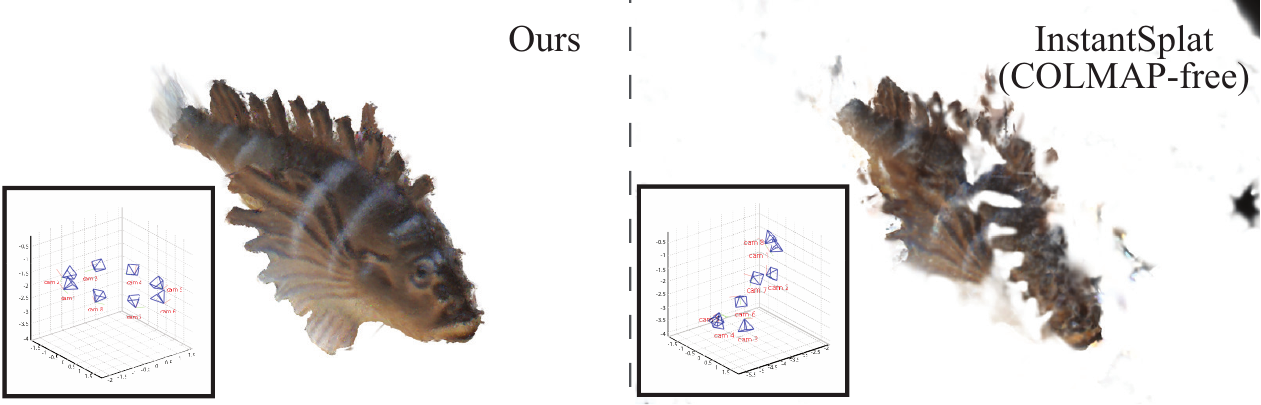}
    \caption{Comparison with a COLMAP-free method~\cite{fan2024instantsplat}.}
    \vspace{-10pt}
    \label{fig:colmap_free}
\end{figure}

Since camera parameters are critical, we also compare our method against a recent COLMAP-free method, InstantSplat~\cite{fan2024instantsplat}. The method bypasses COLMAP by jointly optimizing Gaussian attributes and camera parameters. Visual comparison of a synthesized novel view is shown in Fig.~\ref{fig:colmap_free}. We also show their estimated camera poses and our calibrated poses. Their rendering quality is poor likely because of inaccurate camera pose estimation.  

\section{Conclusion}
\label{sec:conclusion}
We have presented an imaging solution using circularly arranged mirror pairs for full-surround reconstruction and novel view synthesis of miniature scenes. We have thoroughly analyzed the design factors of our mirror lens and derived optimal parameters given the scene size. We have tailored the latest 3DGS framework to allow accurate and robust scene reconstruction using our sparse view input. Our method has been validated through synthetic and real experiments and demonstrated state-of-the-art performance. Since our approach is a single-shot solution, it can be applied to dynamic scenes. One viable direction is to incorporate temporal consistency to allow smooth reconstruction of dynamic scenes.  
\vspace{-4pt}
\section*{Acknowledgments}
This project is partially supported by NSF awards 2225948 and 2238141.
\appendix
%\clearpage
%\setcounter{page}{1}
%\maketitlesupplementary

\section*{Appendices}
 
\section{Geometric Derivations}
\label{sec:derivation}
\subsection{Derivation of Angles in Section 3.2}
\begin{wrapfigure}{r}{0.21\textwidth} 
    \centering
    \vspace{-8pt}
    \includegraphics[width=0.23\textwidth]{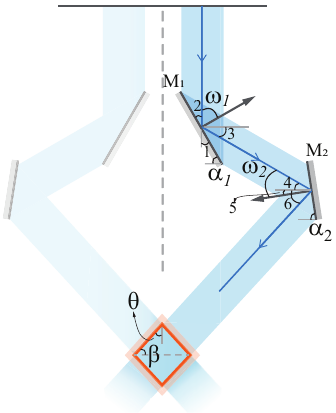}
    \caption{Angle annotations.} 
    \label{fig:ray_geo_sup}
\end{wrapfigure}

Here we show how to derive the angles we used in Section 3.2, when formulating the effective viewing volume. Our goal is derive the half apex angle of the viewing volume ($\theta$), given the tilting angles of the two mirrors ($\alpha_1$ and $\alpha_2$). For ease of reference, we introduce auxiliary angles labeled in numbers. All the angles that we have referred to are annotated in Fig.~\ref{fig:ray_geo_sup}.

Since $\angle 1$ and $\angle 2$ are vertical angles, we have $\angle 2 = \angle 1 = 90^\circ - \alpha_1$. Since $\angle 2$ and $\omega_1$ are complementary, we can calculate the incident/exit angle of reflection on \(M_1\) as:
\begin{equation}
\omega_1 = 90^\circ - \angle 2 = \alpha_1.
\label{eq:omega_1}
\end{equation}

Since $\angle 3$ and $\angle 4$ are alternate angles, we have $\angle 4 = \angle 3 = 2\omega_1 - 90^\circ$. By substituting $\omega_1$ with Eq.~\ref{eq:omega_1}, we have $\angle 4 = 2\alpha_1 - 90^\circ$. Since $\angle 5$ and $\alpha_2$ are complementary, we have $\angle 5 = 90^\circ - \alpha_2$. Therefore, the incident/exit angle of reflection on \(M_2\) can be calculated as:
\begin{equation}
\omega_2 = \angle 4 + \angle 5 = 2\alpha_1 - \alpha_2.
\end{equation}

Since $\angle 6$ and $\beta$ are congruent, we have $\beta = \angle 6 = \angle 5 + \omega_2$. By substituting $\omega_2$ and $\angle 5$, we have $\beta = 90^\circ - 2\Delta\alpha$, where $\Delta\alpha = \alpha_2 - \alpha_1$. The half apex angle of the effective viewing volume, being complenentary to $\beta$, is thus:
\begin{equation}
\label{eq:angle}
\theta = 90^\circ - \beta = 2\Delta\alpha.
\end{equation}

\subsection{Derivation of Conditions in Section 3.2}

\noindent\textbf{Derivation of condition (i).} This condition is introduced to allow light to travel through the lens from one end to the other, after being reflected by the two mirrors in a pair. In addition, the multi-view images formed by the eight mirror pairs should have overlaps, in order to be practical for scene reconstruction. 

\begin{figure}[t]
    \centering
    \includegraphics[width=0.8\linewidth]{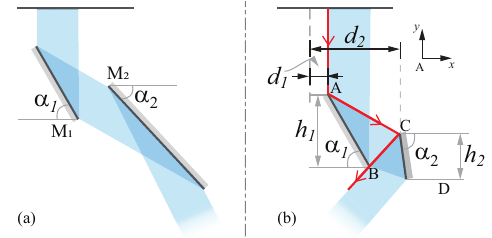}
    \caption{(a) The diverging situation when $\alpha_2<\alpha_1$; (b) The extreme situation without inter-reflection, in which the reflected ray from $M_2$ intersects with the bottom edge of $M_1$.}
    \label{fig:m2_restriction_sup}
\end{figure}

With $\alpha_1 >  45^\circ$ and $\alpha_2 < 90^\circ$, we guarantee that light from the scene could travel through our mirror lens and reach the camera on the other end (\ie, light path wouldn't turn around inside of the lens).  % light path  et's consider a vertical beam light incident to the lens from the camera's end. When $\alpha_1 = 45^\circ$, the vertical beam light would become horizontal after reflection from $M_1$. With that xxx becomes horizontal and xx $\alpha_2 < 90$.
If $\alpha_2 < \alpha_1$, the light exiting the lens (after reflected by $M_2$) would be diverging (see Fig.~\ref{fig:m2_restriction_sup} (a)), resulting none-overlapping multi-view images. So we have $45^\circ < \alpha_1 < \alpha_2 < 90^\circ $.\\

\noindent\textbf{Derivation of condition (ii).} Here we derive the minimum vertically projected height of $M_2$ (denoted as $h_2$), such that it can cover the entire light beam reflected from $M_1$.

The width of parallel light beam reflected from $M_1$ is $w = h_1/ \tan\alpha_1$, where $h_1$ is the vertically projected height of $M_1$. In order to cover the entire beam, the length of $M_2$ (denoted as $l_2$) should satisfy: 
\begin{equation}
l_2 \geq \frac{w}{\sin(\alpha_2 - \angle 4)} = \frac{w}{\cos(2\alpha_1- \alpha_2)}.
\label{eq:l_2}
\end{equation}
Substituting $l_2 = h_2/\sin\alpha_2$ and $w = h_1/ \tan\alpha_1$, we can rewrite Eq.~\ref{eq:l_2} as:
\begin{equation}
h_2 \geq \dfrac{\sin\alpha_2}{\tan\alpha_1 \cdot\cos(2\alpha_1- \alpha_2)} \cdot h_1.
\end{equation}

\noindent\textbf{Derviation of condition (iii).} Here we derive the minimum separation between the two mirrors in order to avoid interreflection. We quantify this distance as $d_2 - d_1$ (where $d_1$ and $d_2$ are the distances from $M_1$ and $M_2$'s upper edges to the central ray), when given their vertically projected heights $h_1$, $h_2$ and tilting angles $\alpha_1$, $\alpha_2$. We consider the extreme situation when the leftmost ray of the light beam intersects with the bottom edge of $M_1$ after reflecting from $M_2$ (see Fig.~\ref{fig:m2_restriction_sup} (b)).

We denote the end points of $M_1$ and $M_2$ in the 2D cross-section plot as $A$, $B$, $C$, and $D$. We setup a coordinate system with \( A \) as the origin as shown in Figure~\ref{fig:m2_restriction_sup}(b). The line equation for $M_1$ (line \( AB \)) can be written as:

\begin{equation}
y = -\tan \alpha_1 \cdot x.
\label{eq:ABline}
\end{equation}
The line equation for the leftmost ray incident to $M_2$ (line \( AC \)) can be written as:
\begin{equation}
y = \cot 2\alpha_1 \cdot x.
\label{eq:ACline}
\end{equation}
Since $x_C = d_2 - d_1$, we plug it into Eq.~\ref{eq:ACline} and calculate the coordinate of \( C \) as $\left( d_2 - d_1, \, \cot 2\alpha_1 \cdot (d_2 - d_1) \right)$. The line equation for leftmost ray reflected from $M_2$ (line \( BC \)) can thus be calculated as:
\begin{equation}
y  = \cot\Delta\alpha \cdot \left( x - (d_2 - d_1) \right)+  \cot 2\alpha_1 \cdot (d_2 - d_1),
\label{eq:BCline}
\end{equation}
where $\Delta\alpha = \alpha_2 - \alpha_1$. By combining Eq.~\ref{eq:ABline} and Eq.~\ref{eq:BCline}, we can calculate the $x$ coordinate of \( B \) as:
\begin{equation}
x_B = \frac{\cot2\Delta\alpha - \cot 2\alpha_1}{\tan \alpha_1 + \cot2\Delta\alpha} \cdot (d_2 - d_1).
\label{eq:x_B}
\end{equation}
To avoid inter-reflection, $x_B$ should be satisfy: $x_B \geq h_1/\tan \alpha_1$. Subsituting $x_B$ with Eq.~\ref{eq:x_B}, we obtain the third condition regarding the mirror distances: 

\begin{equation}
d_2 \geq \dfrac{\tan \alpha_1 + \cot2\Delta\alpha}{\tan \alpha_1\cdot(\cot2\Delta\alpha - \cot 2\alpha_1)} \cdot{h_1} + d_1.
\end{equation}

\section{More Details on Pre-processing Steps}

\begin{figure}[h]
    \centering    
    \includegraphics[width=1\linewidth]{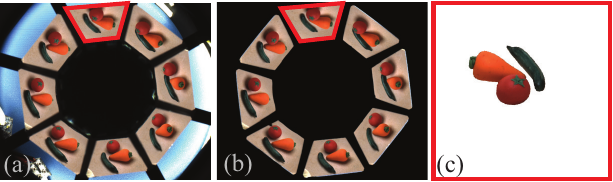}
    \caption{(a) Our captured raw image; (b) Image filtered by the multi-view mask; (c) Re-projected image of the highlighted view.}
    \label{fig:preprocessing}
\end{figure}

Fig.~\ref{fig:preprocessing} shows how our captured raw image is processed into multi-view input to 3DGS. A raw image captured by our portable lens prototype is shown in Fig~\ref{fig:preprocessing} (a). Its resolution is \( 2448 \times 2048 \). We first apply a multi-view mask to extract the effective regions formed through mirror reflection. The filtered image is shown in Fig.~\ref{fig:preprocessing}(b). Then, for each sub-image, we re-project it to allow smooth view transition (we update camera poses after re-projection). We also mask out the background and only reconstruct the foreground objects. The processed image for one sub-view (highlighted in red) is shown in Fig.~\ref{fig:preprocessing} (c). This image is with resolution \( 800 \times 800 \). The eight sub-view images processed in this way are used as input to 3DGS. 

\section{Lens Design Comparison} 
\begin{figure}[h]
    \centering    
    \vspace{-8pt}
    \includegraphics[width=0.9\linewidth]{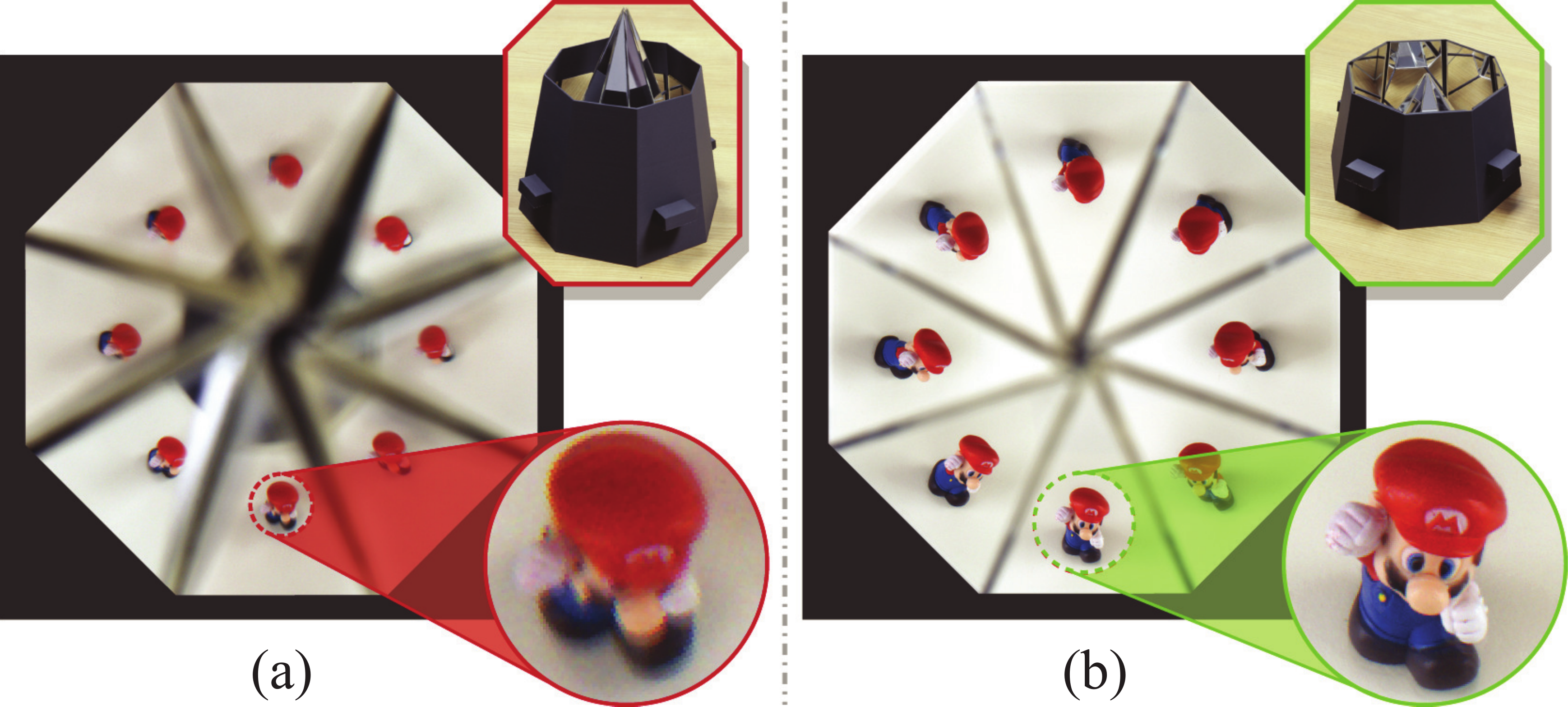}
    \caption{Comparison between two lens designs. Here we show the lens prototypes and their captured images with zoom-in views.}
    \label{fig:optimal_config}
\end{figure}

Here we show comparison between two lens designs with different mirror configuration. Prototypes of the two designs are shown in Fig.~\ref{fig:optimal_config}. The two lenses have the same base lengths for the inner and outer pyramids, with different tilting angles for the mirrors. The parameters we use are shown in Table~\ref{table:lens_para}. Images taken with the two lenses are shown in Fig.~\ref{fig:optimal_config}.

We can see that design (b), which has larger $\Delta\alpha$, has better coverage of the side views (\eg, the figurine's face becomes visible in (b)). This is equivalent to having virtual cameras with more oblique angles. Such configuration is preferred since it provides fuller coverage of the scene. This observation is consistent with our guidelines on optimizing the mirror configuration. 

\begin{table}[h!]
\centering
\caption{Mirror parameters of the two different designs.}
\begin{tabular}{|c|c|c|c|}
\hline
\diaghead(-5, 1){aaaaaaaaaaa}{ }{ }& $\alpha_1$ & $\alpha_2$ & $\Delta\alpha$ \\ \hline
Design (a) & $75^\circ$ & $85^\circ$ & $10^\circ$\\
\hline
Design (b) & $60^\circ$ & $85^\circ$ & $25^\circ$ \\
\hline
\end{tabular}
\label{table:lens_para}
\end{table}

\begin{figure}[b]
    \centering
    \vspace{-8pt}
    \includegraphics[width=1\linewidth]{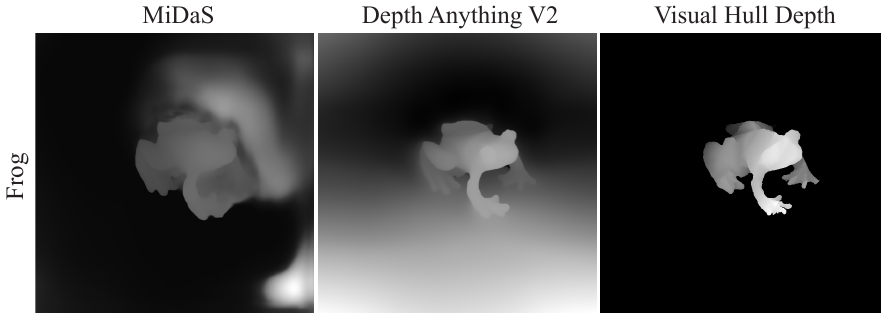}
    \caption{Comparison of depth map obtained by different methods.}
    \label{fig:depth_sup}
\end{figure}

\begin{figure*}[t]
    \centering
    \includegraphics[width=1\linewidth]{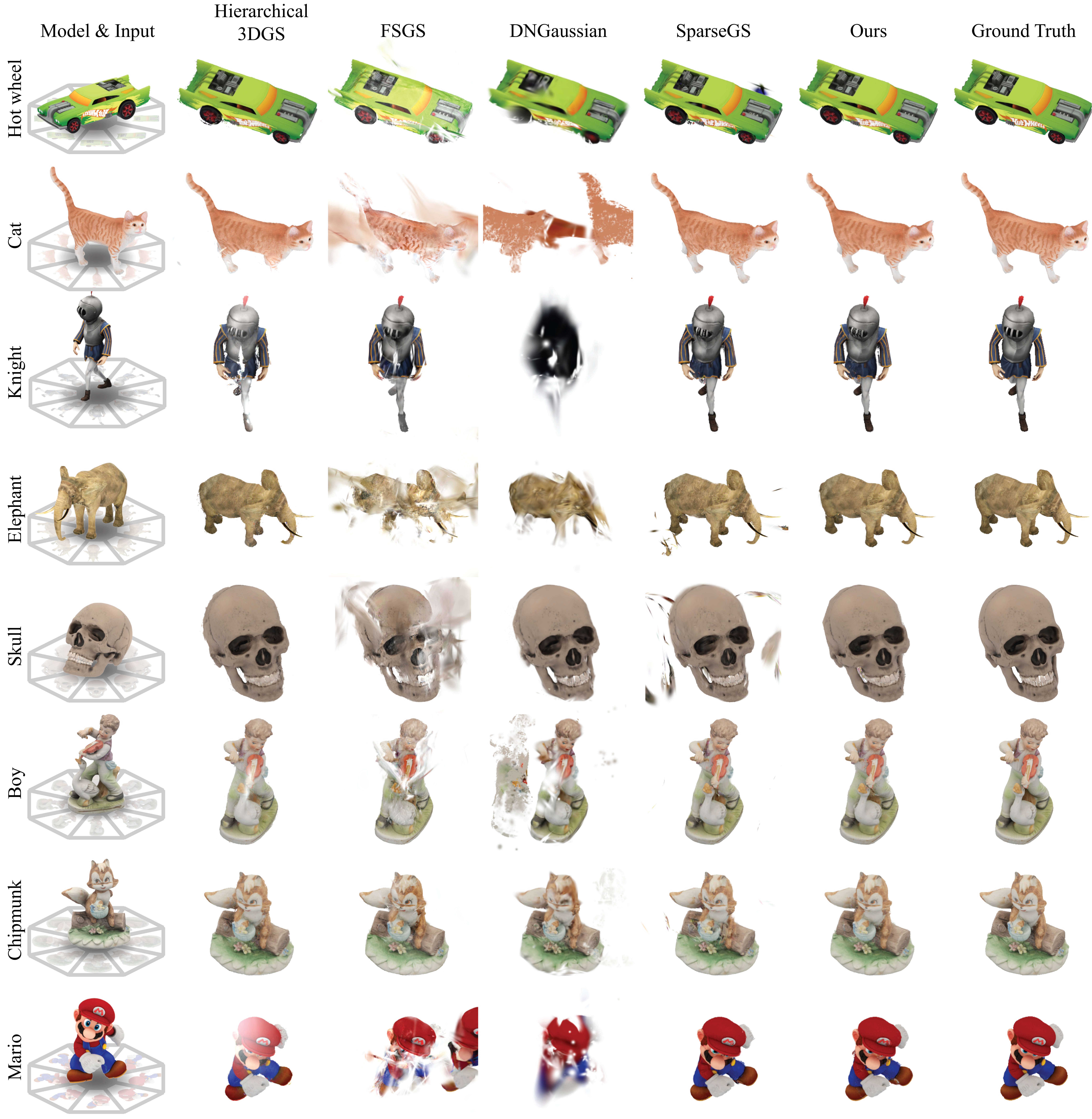}
    \caption{Additional visual comparison results on synthetic data.}
    \label{fig:syn_sup}
\end{figure*}

\begin{figure*}[t]
    \centering
    \includegraphics[width=1\linewidth]{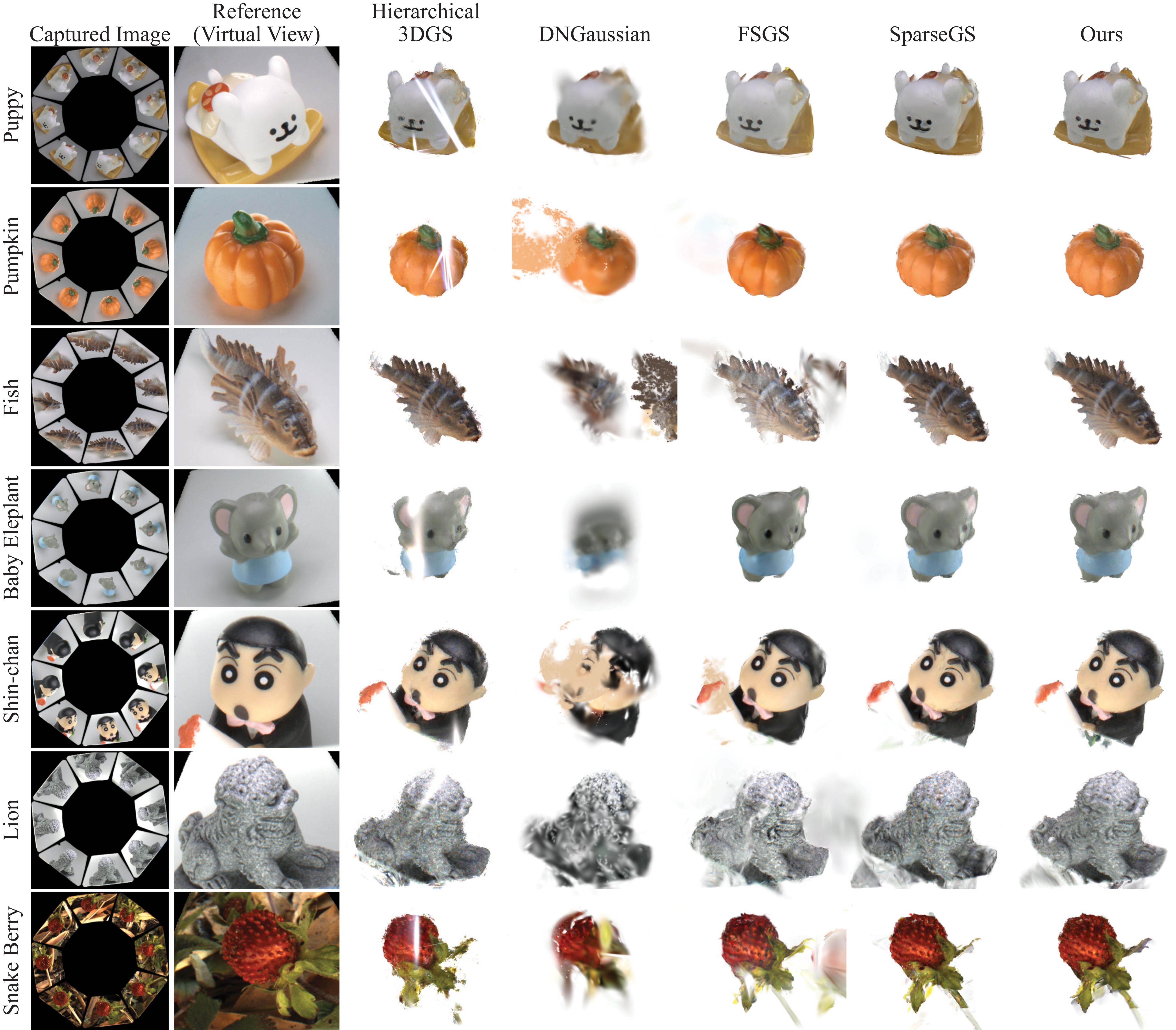}
    \caption{Additional visual comparison results on real data.}
    \label{fig:real_sup}
\end{figure*}

\section{Additional Experimental Results}

\subsection{Ablation on Depth Loss}
Fig.~\ref{fig:depth_sup} compares depth maps obtained by different methods for a real scene (\ie, the ``frog" scene). Specifically, the MiDaS~\cite{Ranftl2022} depth is used by FSGS~\cite{zhu2023FSGS}; Depth Anything V2~\cite{depth_anything_v2} is used by Hierarchical 3DGS~\cite{hierarchicalgaussians24}; and the visual hull depth is used by our approach. We can see that Depth Anything provides much better depth prior than MiDaS depth. Our visual hull depth outperforms Depth Anything result in details (\eg, the frog legs have more discernible depth variation in the visual hull depth). Moreover, the visual hull projection provides depth values in absolute scale, whereas the other two learning-based methods estimate relative depths.

We performed an ablation study on depth loss using the ``skull" scene (see Fig.~\ref{fig:syn_sup}). We compare the PSNR of synthesized novel views for three variants of our algorithm: without depth loss, with monocular depth (Depth Anything V2~\cite{depth_anything_v2} depth), and with visual hull depth (VH depth). The table below shows the ablation study on depth loss. 

\begin{table}[h]
\vspace{-4pt}
\caption{Ablation study on depth loss.}
\centering
\begin{tabular}{c !{\vrule width 0.5pt} c c c}\hline
Variant & w/o depth & w. monodepth & w. VH depth \\
\hline

PSNR & 29.2052 & 29.3698 & 29.3856 \\
\hline
\end{tabular}
\vspace{-8pt}
\label{tab:ablation}
\end{table}

\subsection{Additional Synthetic Results}
We show more visual results on synthetic data in Fig.~\ref{fig:syn_sup}. We compare with recent state-of-the-art 3DGS algorithms: Hierarchical 3DGS~\cite{hierarchicalgaussians24}, FSGS~\cite{zhu2023FSGS}, DNGaussian~\cite{li2024dngaussian}, and SparseGS~\cite{xiong2023sparsegs}. Most of these methods are optimized for spare view input. We can see that our results outperform the state-of-the-arts and resemble the ground truths.

\subsection{Additional Real Results}
Fig.~\ref{fig:real_sup} shows more visual comparison results on real data in comparison with state-of-the-arts. The ``snake berry" scene is captured outdoor with our portable lens. 
\vspace{+26pt}
{
    \small
    \bibliographystyle{ieeenat_fullname}
    \bibliography{main}

\begin{thebibliography}{42}
\providecommand{\natexlab}[1]{#1}
\providecommand{\url}[1]{\texttt{#1}}
\expandafter\ifx\csname urlstyle\endcsname\relax
  \providecommand{\doi}[1]{doi: #1}\else
  \providecommand{\doi}{doi: \begingroup \urlstyle{rm}\Url}\fi

\bibitem[Ahn et~al.(2021)Ahn, Gkioulekas, and
  Sankaranarayanan]{ahn2021kaleidoscopic}
Byeongjoo Ahn, Ioannis Gkioulekas, and Aswin~C Sankaranarayanan.
\newblock Kaleidoscopic structured light.
\newblock \emph{ACM TOG}, 40\penalty0 (6):\penalty0 1--15, 2021.

\bibitem[Ahn et~al.(2023)Ahn, De~Zeeuw, Gkioulekas, and
  Sankaranarayanan]{Ahn2023CVPR}
Byeongjoo Ahn, Michael De~Zeeuw, Ioannis Gkioulekas, and Aswin~C.
  Sankaranarayanan.
\newblock Neural kaleidoscopic space sculpting.
\newblock In \emph{CVPR}, 2023.

\bibitem[Chahl and Srinivasan(1997)]{chahl1997reflective}
Javaan~Singh Chahl and Mandyam~Veerambudi Srinivasan.
\newblock Reflective surfaces for panoramic imaging.
\newblock \emph{Applied optics}, 36\penalty0 (31):\penalty0 8275--8285, 1997.

\bibitem[Davy and Dixon(2019)]{miniature}
Jack Davy and Charlotte Dixon.
\newblock \emph{{Worlds in Miniature}}.
\newblock UCL Press, 2019.

\bibitem[Fan et~al.(2024)Fan, Cong, Wen, Wang, Zhang, Ding, Xu, Ivanovic,
  Pavone, Pavlakos, et~al.]{fan2024instantsplat}
Zhiwen Fan, Wenyan Cong, Kairun Wen, Kevin Wang, Jian Zhang, Xinghao Ding,
  Danfei Xu, Boris Ivanovic, Marco Pavone, Georgios Pavlakos, et~al.
\newblock Instantsplat: Unbounded sparse-view pose-free gaussian splatting in
  40 seconds.
\newblock \emph{arXiv preprint arXiv:2403.20309}, 2, 2024.

\bibitem[Fuchs et~al.(2013)Fuchs, K{\"a}chele, and
  Rusinkiewicz]{Fuchs2013DesignAF}
Martin Fuchs, Markus K{\"a}chele, and Szymon Rusinkiewicz.
\newblock Design and fabrication of faceted mirror arrays for light field
  capture.
\newblock \emph{Comput. Graph. Forum}, 32, 2013.

\bibitem[Gluckman and Nayar(2000)]{Gluckman00CVPR}
Joshua Gluckman and Shree Nayar.
\newblock Rectified catadioptric stereo sensors.
\newblock In \emph{CVPR}, 2000.

\bibitem[Gluckman and Nayar(2001)]{Gluckman2001}
Joshua Gluckman and Shree Nayar.
\newblock Catadioptric stereo using planar mirrors.
\newblock \emph{IJCV}, 44:\penalty0 65--79, 2001.

\bibitem[Han and Perlin(2003)]{Han2003Measuring}
Jefferson~Y. Han and Ken Perlin.
\newblock Measuring bidirectional texture reflectance with a kaleidoscope.
\newblock \emph{ACM TOG}, 22\penalty0 (3), 2003.

\bibitem[Ihrke et~al.(2008)Ihrke, Stich, Gottschlich, Magnor, and
  Seidel]{Ihrke2008FastIL}
Ivo Ihrke, Timo Stich, Heiko Gottschlich, Marcus Magnor, and Hans-Peter Seidel.
\newblock Fast incident light field acquisition and rendering.
\newblock \emph{Journal of {WSCG}}, 16:\penalty0 25--32, 2008.

\bibitem[Ihrke et~al.(2012)Ihrke, Reshetouski, Manakov, Tevs, Wand, and
  Seidel]{Ihrke2012}
Ivo Ihrke, Ilya Reshetouski, Alkhazur Manakov, Art Tevs, Michael Wand, and
  Hans-Peter Seidel.
\newblock A kaleidoscopic approach to surround geometry and reflectance
  acquisition.
\newblock In \emph{CVPRW}, 2012.

\bibitem[Kerbl et~al.(2023)Kerbl, Kopanas, Leimk{\"u}hler, and
  Drettakis]{kerbl3Dgaussians}
Bernhard Kerbl, Georgios Kopanas, Thomas Leimk{\"u}hler, and George Drettakis.
\newblock 3{D} gaussian splatting for real-time radiance field rendering.
\newblock \emph{ACM TOG}, 42\penalty0 (4), 2023.

\bibitem[Kerbl et~al.(2024)Kerbl, Meuleman, Kopanas, Wimmer, Lanvin, and
  Drettakis]{hierarchicalgaussians24}
Bernhard Kerbl, Andreas Meuleman, Georgios Kopanas, Michael Wimmer, Alexandre
  Lanvin, and George Drettakis.
\newblock A hierarchical 3d gaussian representation for real-time rendering of
  very large datasets.
\newblock \emph{ACM TOG}, 43\penalty0 (4), 2024.

\bibitem[Kim et~al.(2006)Kim, Yoon, Kim, and Kweon]{Kim2006}
Jungho Kim, Kuk-jin Yoon, Jun-sik Kim, and Inso Kweon.
\newblock Visual slam by single-camera catadioptric stereo.
\newblock In \emph{SICE-ICASE International Joint Conference}, 2006.

\bibitem[Kirillov et~al.(2023)Kirillov, Mintun, Ravi, Mao, Rolland, Gustafson,
  Xiao, Whitehead, Berg, Lo, Doll{\'a}r, and Girshick]{kirillov2023segany}
Alexander Kirillov, Eric Mintun, Nikhila Ravi, Hanzi Mao, Chloe Rolland, Laura
  Gustafson, Tete Xiao, Spencer Whitehead, Alexander~C. Berg, Wan-Yen Lo, Piotr
  Doll{\'a}r, and Ross Girshick.
\newblock Segment anything.
\newblock \emph{arXiv:2304.02643}, 2023.

\bibitem[Kocabas et~al.(2024)Kocabas, Chang, Gabriel, Tuzel, and
  Ranjan]{kocabas2024hugs}
Muhammed Kocabas, Jen-Hao~Rick Chang, James Gabriel, Oncel Tuzel, and Anurag
  Ranjan.
\newblock Hugs: Human gaussian splats.
\newblock In \emph{CVPR}, pages 505--515, 2024.

\bibitem[Lanman et~al.(2007)Lanman, Crispell, and Taubin]{Lanman2007}
Douglas Lanman, Daniel Crispell, and Gabriel Taubin.
\newblock Surround structured lighting for full object scanning.
\newblock In \emph{International Conference on 3-D Digital Imaging and Modeling
  (3DIM)}, 2007.

\bibitem[Laurentini(1994)]{laurentini1994visual}
Aldo Laurentini.
\newblock The visual hull concept for silhouette-based image understanding.
\newblock \emph{IEEE TPAMI}, 16\penalty0 (2):\penalty0 150--162, 1994.

\bibitem[Li et~al.(2024)Li, Zhang, Bai, Zheng, Ning, Zhou, and
  Gu]{li2024dngaussian}
Jiahe Li, Jiawei Zhang, Xiao Bai, Jin Zheng, Xin Ning, Jun Zhou, and Lin Gu.
\newblock Dngaussian: Optimizing sparse-view 3d gaussian radiance fields with
  global-local depth normalization.
\newblock In \emph{CVPR}, 2024.

\bibitem[Mas et~al.(2019)Mas, Druart, Vach{\'e}, Favier, Alazarine, Compain,
  Morin, and Gu{\'e}rineau]{mas2019kaleidoscope}
Adrien Mas, Guillaume Druart, Maxime Vach{\'e}, Sylvain Favier, Aymeric
  Alazarine, Eric Compain, Nathalie Morin, and Nicolas Gu{\'e}rineau.
\newblock Kaleidoscope-based multi-view infrared system.
\newblock \emph{Optics Letters}, 44\penalty0 (20):\penalty0 4977--4980, 2019.

\bibitem[Mildenhall et~al.(2020)Mildenhall, Srinivasan, Tancik, Barron,
  Ramamoorthi, and Ng]{mildenhall2020nerf}
Ben Mildenhall, Pratul~P. Srinivasan, Matthew Tancik, Jonathan~T. Barron, Ravi
  Ramamoorthi, and Ren Ng.
\newblock Nerf: Representing scenes as neural radiance fields for view
  synthesis.
\newblock In \emph{ECCV}, 2020.

\bibitem[Mukaigawa et~al.(2011)Mukaigawa, Tagawa, Kim, Raskar, Matsushita, and
  Yagi]{mukaigawa2011hemispherical}
Yasuhiro Mukaigawa, Seiichi Tagawa, Jaewon Kim, Ramesh Raskar, Yasuyuki
  Matsushita, and Yasushi Yagi.
\newblock Hemispherical confocal imaging using turtleback reflector.
\newblock In \emph{ACCV}, 2011.

\bibitem[Nene and Nayar(1998)]{Nene1998StereoWM}
Sameer~A. Nene and Shree~K. Nayar.
\newblock Stereo with mirrors.
\newblock \emph{ICCV}, pages 1087--1094, 1998.

\bibitem[Ranftl et~al.(2022)Ranftl, Lasinger, Hafner, Schindler, and
  Koltun]{Ranftl2022}
Ren\'{e} Ranftl, Katrin Lasinger, David Hafner, Konrad Schindler, and Vladlen
  Koltun.
\newblock Towards robust monocular depth estimation: Mixing datasets for
  zero-shot cross-dataset transfer.
\newblock \emph{IEEE TPAMI}, 44\penalty0 (3), 2022.

\bibitem[Reshetouski and Ihrke(2013)]{reshetouski2013mirrors}
Ilya Reshetouski and Ivo Ihrke.
\newblock \emph{Mirrors in computer graphics, computer vision and
  time-of-flight imaging}.
\newblock 2013.

\bibitem[Reshetouski et~al.(2011)Reshetouski, Manakov, Seidel, and
  Ihrke]{reshetouski2011three}
Ilya Reshetouski, Alkhazur Manakov, Hans-Peter Seidel, and Ivo Ihrke.
\newblock Three-dimensional kaleidoscopic imaging.
\newblock In \emph{CVPR}, 2011.

\bibitem[Sch\"{o}nberger and Frahm(2016)]{schoenberger2016sfm}
Johannes~Lutz Sch\"{o}nberger and Jan-Michael Frahm.
\newblock Structure-from-motion revisited.
\newblock In \emph{CVPR}, 2016.

\bibitem[Svoboda and Pajdla(2002)]{svoboda2002epipolar}
Tom{\'a}{\v{s}} Svoboda and Tom{\'a}{\v{s}} Pajdla.
\newblock Epipolar geometry for central catadioptric cameras.
\newblock \emph{IJCV}, 49:\penalty0 23--37, 2002.

\bibitem[Swaminathan et~al.(2003)Swaminathan, Nayar, and
  Grossberg]{swaminathan2003framework}
Rahul Swaminathan, Shree~K Nayar, and Michael~D Grossberg.
\newblock Framework for designing catadioptric projection and imaging systems.
\newblock \emph{ICCV}, 2003.

\bibitem[Tang et~al.(2024)Tang, Ren, Zhou, Liu, and
  Zeng]{tang2023dreamgaussian}
Jiaxiang Tang, Jiawei Ren, Hang Zhou, Ziwei Liu, and Gang Zeng.
\newblock Dreamgaussian: Generative gaussian splatting for efficient 3d content
  creation.
\newblock \emph{ICLR}, 2024.

\bibitem[Tremblay et~al.(2007)Tremblay, Stack, Morrison, and Ford]{Tremblay07}
Eric~J. Tremblay, Ronald~A. Stack, Rick~L. Morrison, and Joseph~E. Ford.
\newblock Ultrathin cameras using annular folded optics.
\newblock \emph{Appl. Opt.}, 46\penalty0 (4):\penalty0 463--471, 2007.

\bibitem[Tremblay et~al.(2009)Tremblay, Stack, Morrison, Karp, and
  Ford]{Tremblay2009UltrathinFI}
Eric~J. Tremblay, Ronald~A. Stack, Rick~L. Morrison, Jason~Harris Karp, and
  Joseph~E. Ford.
\newblock Ultrathin four-reflection imager.
\newblock \emph{Applied optics}, 48 2:\penalty0 343--54, 2009.

\bibitem[Wechsler et~al.(2022)Wechsler, Heintzmann, and
  Ihrke]{wechsler2022kaleidomicroscope}
Felix Wechsler, Rainer Heintzmann, and Ivo Ihrke.
\newblock Kaleidomicroscope-a kaleidoscopic multiview microscope.
\newblock In \emph{Computational Optical Sensing and Imaging}, pages CTu4F--5,
  2022.

\bibitem[Wolf et~al.(2024)Wolf, Bracha, and Kimmel]{wolf2024gs2mesh}
Yaniv Wolf, Amit Bracha, and Ron Kimmel.
\newblock Gs2mesh: Surface reconstruction from gaussian splatting via novel
  stereo views.
\newblock In \emph{ECCV}, 2024.

\bibitem[Wu and Chang(2010)]{Wu2010DesignOS}
Hsien-Huang~P. Wu and Shih-Hsin Chang.
\newblock Design of stereoscopic viewing system based on a compact mirror and
  dual monitor.
\newblock \emph{Optical Engineering}, 49:\penalty0 027401, 2010.

\bibitem[Wu et~al.(2024)Wu, Mildenhall, Henzler, Park, Gao, Watson, Srinivasan,
  Verbin, Barron, Poole, and Holynski]{wu2024reconfusion}
Rundi Wu, Ben Mildenhall, Philipp Henzler, Keunhong Park, Ruiqi Gao, Daniel
  Watson, Pratul~P Srinivasan, Dor Verbin, Jonathan~T Barron, Ben Poole, and
  Aleksander Holynski.
\newblock Reconfusion: 3d reconstruction with diffusion priors.
\newblock In \emph{CVPR}, 2024.

\bibitem[Xiong et~al.(2023)Xiong, Muttukuru, Upadhyay, Chari, and
  Kadambi]{xiong2023sparsegs}
Haolin Xiong, Sairisheek Muttukuru, Rishi Upadhyay, Pradyumna Chari, and Achuta
  Kadambi.
\newblock {SparseGS}: Real-time 360° sparse view synthesis using gaussian
  splatting.
\newblock \emph{Arxiv}, 2023.

\bibitem[Yang et~al.(2024{\natexlab{a}})Yang, Li, Fang, Liang, Xie, Zhang,
  Shen, and Tian]{yang2024gaussianobject}
Chen Yang, Sikuang Li, Jiemin Fang, Ruofan Liang, Lingxi Xie, Xiaopeng Zhang,
  Wei Shen, and Qi Tian.
\newblock Gaussianobject: High-quality 3d object reconstruction from four views
  with gaussian splatting.
\newblock \emph{ACM TOG}, 43\penalty0 (6), 2024{\natexlab{a}}.

\bibitem[Yang et~al.(2024{\natexlab{b}})Yang, Kang, Huang, Zhao, Xu, Feng, and
  Zhao]{depth_anything_v2}
Lihe Yang, Bingyi Kang, Zilong Huang, Zhen Zhao, Xiaogang Xu, Jiashi Feng, and
  Hengshuang Zhao.
\newblock Depth anything v2.
\newblock \emph{arXiv:2406.09414}, 2024{\natexlab{b}}.

\bibitem[Zhang et~al.(2024)Zhang, Fang, Shrestha, Liang, Long, and
  Tan]{zhang2024rade}
Baowen Zhang, Chuan Fang, Rakesh Shrestha, Yixun Liang, Xiaoxiao Long, and Ping
  Tan.
\newblock Rade-gs: Rasterizing depth in gaussian splatting.
\newblock \emph{arXiv preprint arXiv:2406.01467}, 2024.

\bibitem[Zhou et~al.(2024)Zhou, Lin, Shan, Wang, Sun, and
  Yang]{zhou2024drivinggaussian}
Xiaoyu Zhou, Zhiwei Lin, Xiaojun Shan, Yongtao Wang, Deqing Sun, and Ming-Hsuan
  Yang.
\newblock Drivinggaussian: Composite gaussian splatting for surrounding dynamic
  autonomous driving scenes.
\newblock In \emph{CVPR}, pages 21634--21643, 2024.

\bibitem[Zhu et~al.(2024)Zhu, Fan, Jiang, and Wang]{zhu2023FSGS}
Zehao Zhu, Zhiwen Fan, Yifan Jiang, and Zhangyang Wang.
\newblock {FSGS}: Real-time few-shot view synthesis using gaussian splatting.
\newblock In \emph{ECCV}, 2024.

\end{thebibliography}


\begin{thebibliography}{6}
\providecommand{\natexlab}[1]{#1}
\providecommand{\url}[1]{\texttt{#1}}
\expandafter\ifx\csname urlstyle\endcsname\relax
  \providecommand{\doi}[1]{doi: #1}\else
  \providecommand{\doi}{doi: \begingroup \urlstyle{rm}\Url}\fi

\bibitem[Kerbl et~al.(2024)Kerbl, Meuleman, Kopanas, Wimmer, Lanvin, and
  Drettakis]{hierarchicalgaussians24}
Bernhard Kerbl, Andreas Meuleman, Georgios Kopanas, Michael Wimmer, Alexandre
  Lanvin, and George Drettakis.
\newblock A hierarchical 3d gaussian representation for real-time rendering of
  very large datasets.
\newblock \emph{ACM TOG}, 43\penalty0 (4), 2024.

\bibitem[Li et~al.(2024)Li, Zhang, Bai, Zheng, Ning, Zhou, and
  Gu]{li2024dngaussian}
Jiahe Li, Jiawei Zhang, Xiao Bai, Jin Zheng, Xin Ning, Jun Zhou, and Lin Gu.
\newblock Dngaussian: Optimizing sparse-view 3d gaussian radiance fields with
  global-local depth normalization.
\newblock In \emph{CVPR}, 2024.

\bibitem[Ranftl et~al.(2022)Ranftl, Lasinger, Hafner, Schindler, and
  Koltun]{Ranftl2022}
Ren\'{e} Ranftl, Katrin Lasinger, David Hafner, Konrad Schindler, and Vladlen
  Koltun.
\newblock Towards robust monocular depth estimation: Mixing datasets for
  zero-shot cross-dataset transfer.
\newblock \emph{IEEE TPAMI}, 44\penalty0 (3), 2022.

\bibitem[Xiong et~al.(2023)Xiong, Muttukuru, Upadhyay, Chari, and
  Kadambi]{xiong2023sparsegs}
Haolin Xiong, Sairisheek Muttukuru, Rishi Upadhyay, Pradyumna Chari, and Achuta
  Kadambi.
\newblock {SparseGS}: Real-time 360° sparse view synthesis using gaussian
  splatting.
\newblock \emph{Arxiv}, 2023.

\bibitem[Yang et~al.(2024)Yang, Kang, Huang, Zhao, Xu, Feng, and
  Zhao]{depth_anything_v2}
Lihe Yang, Bingyi Kang, Zilong Huang, Zhen Zhao, Xiaogang Xu, Jiashi Feng, and
  Hengshuang Zhao.
\newblock Depth anything v2.
\newblock \emph{arXiv:2406.09414}, 2024.

\bibitem[Zhu et~al.(2024)Zhu, Fan, Jiang, and Wang]{zhu2023FSGS}
Zehao Zhu, Zhiwen Fan, Yifan Jiang, and Zhangyang Wang.
\newblock {FSGS}: Real-time few-shot view synthesis using gaussian splatting.
\newblock In \emph{ECCV}, 2024.

\end{thebibliography}
}


\end{document}